\documentclass[letterpaper]{article} 
\usepackage{aaai2026}
\usepackage{times}  
\usepackage{helvet}  
\usepackage{courier}  
\usepackage[hyphens]{url}  
\usepackage{graphicx} 
\usepackage{natbib}  
\usepackage{caption} 
\usepackage{algorithm}
\usepackage{algorithmic}

\usepackage{newfloat}
\usepackage{listings}
\DeclareCaptionStyle{ruled}{labelfont=normalfont,labelsep=colon,strut=off} 
\floatstyle{ruled}
\newfloat{listing}{tb}{lst}{}
\floatname{listing}{Listing}

\usepackage{booktabs}
\usepackage{multirow}
\usepackage{makecell}
\usepackage{amsmath}
\usepackage{subcaption}
\usepackage{xcolor}
\newcommand{\yes}{\textcolor{blue}{yes}}
\newcommand{\no}{\textcolor{blue}{no}}
\title{How Effective are Large Time Series Models in Hydrology? \\ A Study on Water Level Forecasting in Everglades}
\author {
    Rahuul Rangaraj\textsuperscript{\rm 1}\equalcontrib,
    Jimeng Shi\textsuperscript{\rm 1}\equalcontrib, \\
    Azam Shirali\textsuperscript{\rm 1},
    Rajendra Paudel\textsuperscript{\rm 2},
    Yanzhao Wu\textsuperscript{\rm 1},
    Giri Narasimhan\textsuperscript{\rm 1}
}
\affiliations {
    \textsuperscript{\rm 1}Florida International University, 
    \textsuperscript{\rm 2}National Park Service, U.S.\\
    \texttt{\{rrang016,jshi008,ashir018,yawu,giri\}@fiu.edu, rajendra\_paudel@nps.gov}
}

\usepackage{bibentry}

\begin{document}

\maketitle

\begin{abstract}
The Everglades play a crucial role in flood and drought regulation, water resource planning, and ecosystem management in the surrounding regions.
However, traditional physics-based and statistical methods for predicting water levels often face significant challenges, including high computational costs and limited adaptability to diverse or unforeseen conditions.
Recent advancements in large time series models have demonstrated the potential to address these limitations, with state-of-the-art deep learning and foundation models achieving remarkable success in time series forecasting across various domains.
Despite this progress, their application to critical environmental systems, such as the Everglades, remains underexplored.
In this study, we fill the gap by investigating twelve \emph{task-specific} models and five \emph{time series foundation models} across six categories for a real-world application focused on water level prediction in the Everglades. 
Our primary results show that the foundation model \texttt{Chronos} significantly outperforms all other models while the remaining foundation models exhibit relatively poor performance.
We also noticed that the performance of task-specific models varies with the model architectures, and discussed the possible reasons.
We hope our study and findings will inspire the community to explore the applicability of large time series models in hydrological applications.
The code and data are available at \textcolor{brown}{\url{https://github.com/rahuul2992000/Everglades-Benchmark}}.
\end{abstract}

\section{Introduction}
\begin{figure}[ht]
\centering
    \includegraphics[width=0.95\columnwidth]{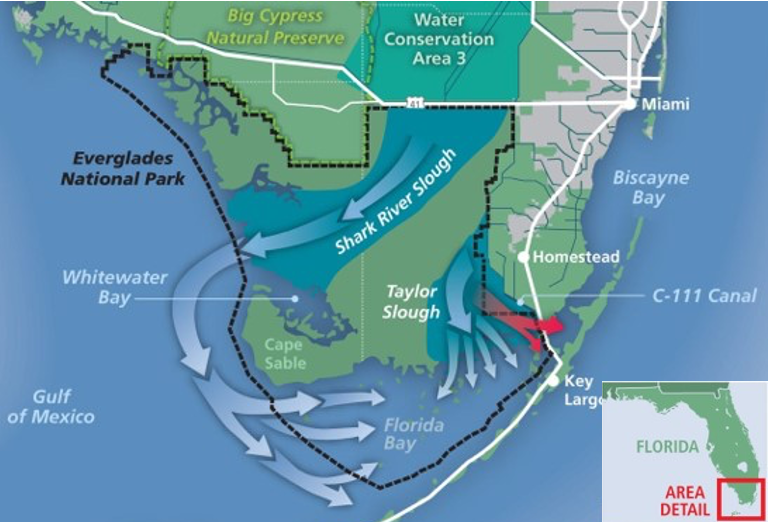}
\caption{Water Flow through Everglades National Park.}
\label{fig:everglade_domain}
\vspace{-3mm}
\end{figure}

The Everglades is a unique ecosystem characterized by vast, swampy grassland, especially containing sawgrass, and seasonally covered by slowly moving water. Water levels in the Everglades play a pivotal role in the prevention of floods and droughts, sustainable water resource planning, recreational and infrastructure management, and ecosystem management \cite{paudel2020assessing,saberski2022improved}.
For example, the U.S. Everglades National Park in South Florida, spanning 2,357 square miles, as shown in Figure~\ref{fig:everglade_domain}, has a profound influence on urban development, flood control, agricultural expansion, and natural hydrology. 
Therefore, accurate water level forecasting in the Everglades is vital, promoting environmental sustainability and benefiting society. 

Water levels are typically recorded as time series, and recent advancements in deep learning (DL) have demonstrated its significant potential in time series forecasting. 
These models excel at capturing nonlinear dependencies across time and variables, enabling both short- and long-term predictions~\cite{kumar2023state}.
For example, recurrent neural networks (RNNs) and convolutional neural networks (CNNs) have been used to learn the temporal dependencies for the prediction of water level~\cite{shi2023deep}.
Graph neural networks, often integrated with sequential models, have been utilized to learn complex spatiotemporal relationships from multiple environmental variables~\cite{kazadi2022flood}..
More broadly, state-of-the-art models for time series forecasting have demonstrated remarkable performance across diverse domains~\cite{liu2024deep}, such as electricity, weather, traffic, healthcare, and energy. Most of them leverage Transformer-based architectures~\cite{vaswani2017attention}, which excel at capturing long-term correlations among elements within time series.
However, these models usually require high computation costs due to the self-attention scheme with $\mathcal{O}(L^2)$ time and memory complexity, where $L$ is the sequence length. 
Although several variants of Transformers~\cite{li2019enhancing,zhou2021informer,nie2022time} have improved the computational complexity to $\mathcal{O}(L\log L)$ or $\mathcal{O}(L)$, significant challenges still persist.
Furthermore, some researchers have questioned whether the self-attention mechanism captures temporal information, claiming that even linear models with $\mathcal{O}(L)$ complexity can outperform Transformer-based models for long-term time series forecasting~\cite{zeng2023transformers}. 
Additionally, some researchers have attempted to leverage large language models (LLMs) for time series forecasting~\cite{jin2023time}. 
These efforts focus on adapting the comprehensive prior knowledge of LLMs to the time series domain~\cite{jin2024position}.  
While these models have demonstrated exceptional performance, they are fully \emph{task-specific}, requiring retraining from scratch for each new dataset or experimental setting.

Inspired by the success of large language models (LLMs) in natural language processing, researchers have explored the development of \emph{foundation models} tailored to process time series data.
These models adopt a training paradigm similar to that used for LLMs, where each time point in a time series is treated as analogous to a ``word'' or ``token'' in a sentence.
By leveraging the self-supervised learning technique, these models are pre-trained on large-scale time series datasets to capture complex temporal patterns. The resulting pre-trained models can be fine-tuned to adapt to various downstream tasks.
The first time series foundation model is called \texttt{TimeGPT}~\cite{garza2023timegpt}. Other examples include \texttt{TimesFM}~\cite{das2023decoder}, 
\texttt{Timer}~\cite{liu2024timer}, \texttt{Moirai}~\cite{woo2024unified} and \texttt{Chronos}~\cite{ansari2024chronos}, etc.
They have demonstrated remarkable performance in a zero-shot setting, reducing or eliminating the need for task-specific retraining.

While these task-specific and foundation models have shown unparalleled performance in time series forecasting across various domains, their application to hydrology is largely unexplored.
We study \emph{the applicability of these models for water level prediction in the Everglades}, with profound implications in hydrology. 
Our contributions are:
\begin{itemize}
    \item We evaluated 12 task-specific models and 5 time-series foundation models on the accuracy and efficiency of water level prediction in the Everglades.
    \item We conducted statistical analyses of prediction errors by assessing models in worst-performing cases, particularly for extremely high and low water levels. 
    \item We examined the unique features of foundation models, which support varying input lengths and outputs without requiring retraining. 
\end{itemize}

\section{Methodology}

\subsection{Region Selection and Data Collection}
\begin{figure}[ht!]
\centering
    \includegraphics[width=0.95\columnwidth]{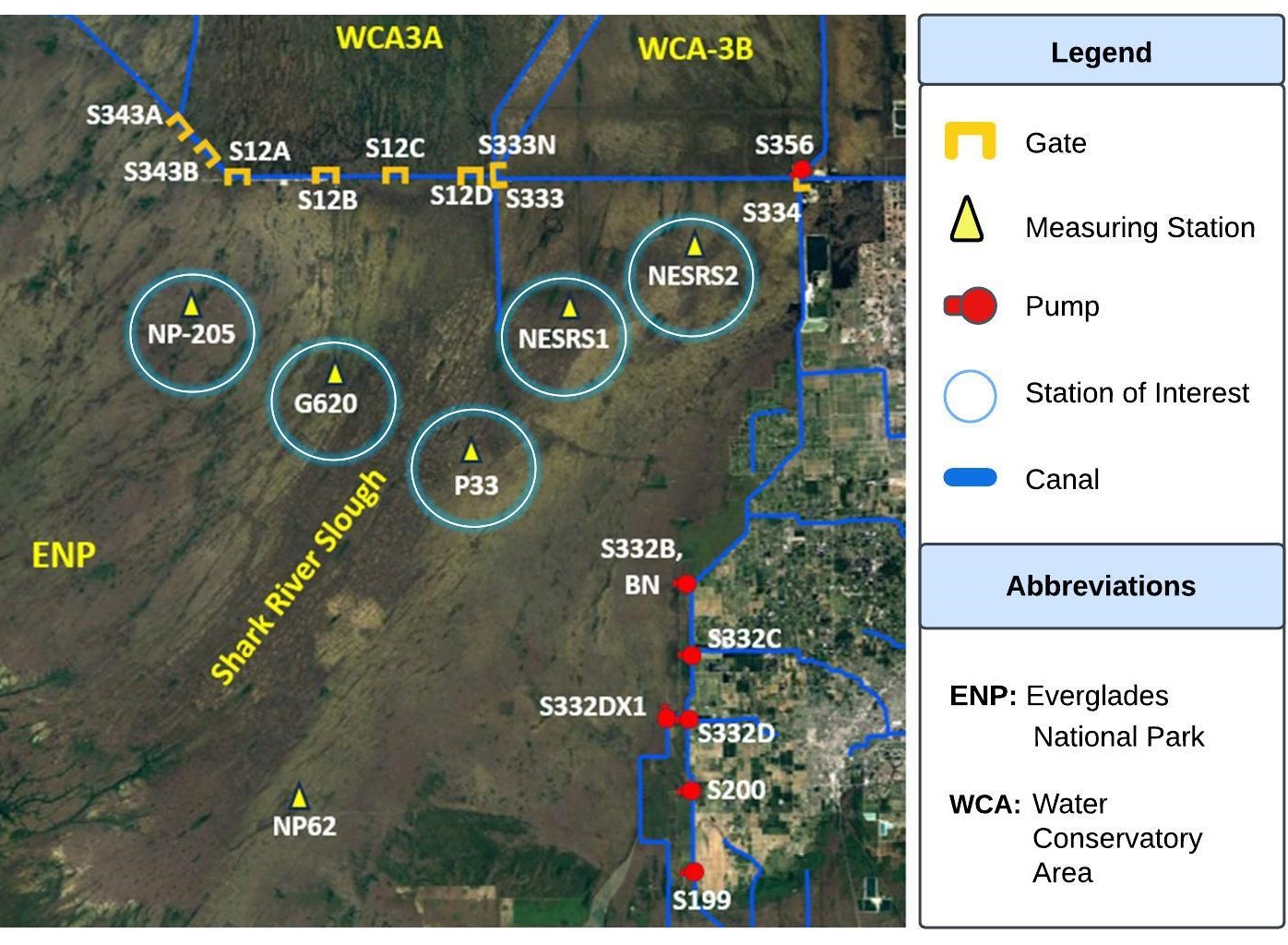}
\caption{Study area: a portion of the Everglades National Park, covering key hydrological structures and stations.}
\label{fig:everglade_map}
\end{figure}

The study domain focuses on the Everglades National Park (see Figure \ref{fig:everglade_map}), including various hydrological structures and stations, and measurements, such as water levels, water flows, precipitation, and evapotranspiration values. 
We downloaded the corresponding data from the Everglades Depth Estimation Network (EDEN) \cite{haider2020everglades} and South Florida Water Management District's Environmental Database (DBHYDRO) \cite{sfwmd2024}. 
The selection process resulted in a dataset with 37 variables spanning daily records of water levels, water flows, rainfall, and potential evapotranspiration (PET). 
We provide a summary of the data in Appendix \ref{sec:dataset}.

\begin{figure*}[htbp]
\centering
    \includegraphics[width=\textwidth]{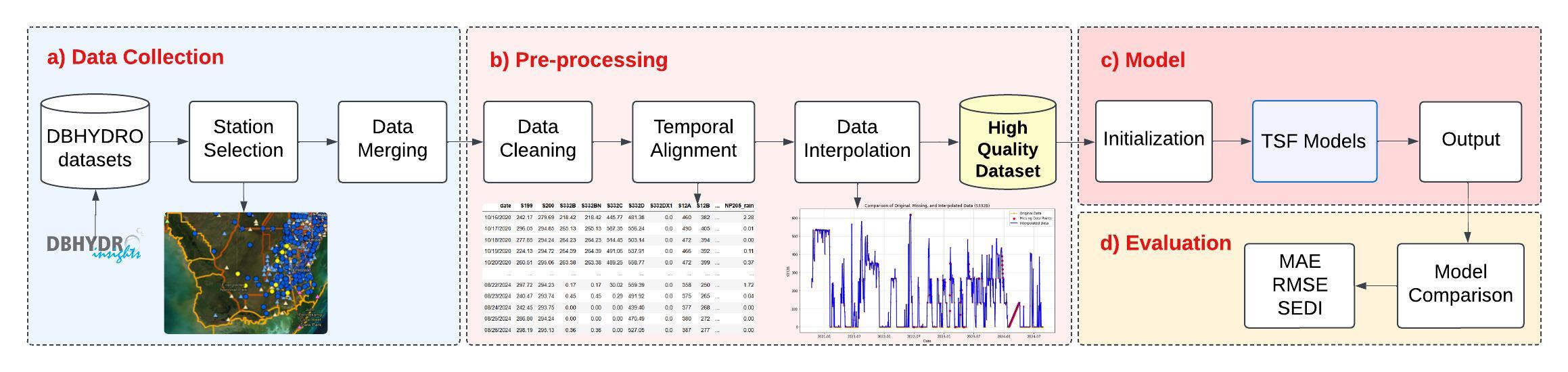}
\caption{Flow diagram of the Methodology. a) Data collection: retrieval and merging of raw data. b) Preprocessing: data cleaning, temporal alignment, and interpolation. c) Time-series forecasting (TSF) models (see Section \ref{sec:models}). d) Evaluation with MAE, RMSE, and SEDI metrics.
Details on data collection and pre-processing are provided in Appendix \ref{sec:dataset}.
}
\label{fig:Methodology}
\end{figure*}

\subsection{State-of-The-Art Models}
\label{sec:models}
This section briefly introduces task-specific and foundation models for time series forecasting.

\subsubsection{Linear-based Models}
\texttt{DLinear} and \texttt{NLinear}, introduced in~\cite{zeng2023transformers}, are two simple yet effective linear models designed for long-term time series forecasting. \texttt{DLinear} decomposes the raw input data into two components: a trend component, extracted using a moving average kernel, and a remainder (seasonal) component. Each component is processed independently by a single-layer linear model to learn its feature representation, and the two components are then summed up to generate the final prediction.
\texttt{NLinear} operates by first subtracting the last value of the input sequence from all elements in the sequence. The adjusted input is then passed through a linear layer to learn the feature representation, after which the subtracted value is added back to generate the final prediction.

\subsubsection{MLP-based Models}
\texttt{NBEATS}~\cite{oreshkin2019n} employs a deep neural architecture based on backward and forward residual links and a deep stack of fully-connected layers.
The stacked architecture comprises multiple fully connected blocks, enabling iterative refinement of predictions through backward and forward forecasting.
\texttt{TimeMixer}~\cite{wang2024timemixer} is based on multi-layer perceptrons (MLPs) and designed for multi-scale time series modeling, capturing both microscopic seasonal and macroscopic trends.
\texttt{TSMixer} is a novel architecture comprising stacked MLPs, which learns from both time and feature dimensions with mixing operations.
\texttt{TSMixerx}~\cite{chen2023tsmixer} is an extension of \texttt{TSMixer}, allowing additional exogenous features, including static features and future information, such as product categories and promotional events.

\subsubsection{Transformer-based Models}
\indent \texttt{PatchTST}~\cite{nie2022time} segments long time series into multiple non-overlapping patches of fixed length, aggregating time steps into subseries-level patches to capture rich local information that is otherwise unavailable at the point level. 
Additionally, it leverages the Transformer architecture with self-attention mechanisms to model temporal dependencies across these patches, enabling the capture of long-term dependencies while reducing computational overhead.
\texttt{Informer}~\cite{zhou2021informer} addresses the computational challenges of vanilla Transformers in long-horizon time series forecasting by introducing three key innovations. First, it employs a self-attention mechanism, which reduces the time and memory complexity from $\mathcal{O}(L^2)$ to $\mathcal{O}(L\log L)$ by focusing on the most critical query-key pairs. Second, it incorporates a self-attention distillation process, enabling the model to retain essential temporal dependencies while effectively handling long input sequences. 
\texttt{iTransformer}~\cite{liu2023itransformer} modifies the traditional Transformer architecture by implementing the \textit{attention} mechanism across multiple variables rather than individual time points, where each time series is embedded as a token. 
This model has shown promising performance in capturing cross-variable dependencies in multivariate time series data.

\subsubsection{KAN-based Models}
\texttt{KANs} (Kolmogorov-Arnold Networks)~\cite{liu2024kan} are an alternative to MLPs, using the Kolmogorov-Arnold approximation theorem, where splines are learned to approximate more complex functions. Unlike the MLP, the nonlinear functions are learned at the edges. At the nodes, the edge functions are merely summed up.
\texttt{RMoK} (Reversible Mixture of KAN)~\cite{han2024kan4tsf} builds on the KAN concept and employs a mixture-of-experts approach to capture both global and local patterns in time series data. 
It uses a gating network to assign variables to specialized KAN experts, such as Wav-KAN, JacobiKAN, and TaylorKAN. These experts utilize different mathematical functions and polynomials to effectively model complex time series relationships.

\subsubsection{LLM-based Models}
\texttt{TimeLLM}~\cite{jin2023time} is a reprogramming framework to repurpose LLMs without changes for general time series forecasting. 
First, it reprograms the input time series with text prototypes by Prompt-as-Prefix (PaP), and then both time series and prompts are fed into a native LLM to align the two modalities. 
In other words, \texttt{TimeLLM} transforms a forecasting task into a ``language task'' that can be tackled by an off-the-shelf LLM.

\subsubsection{Foundation Models}
\texttt{TimeGPT}~\cite{garza2023timegpt} is the first foundation model pre-trained with enormous amounts of time series data. It is capable of generating accurate zero-shot inference for diverse datasets not seen during training. It is engineered to process time series of different characteristics while accommodating different input sizes and forecasting horizons. 
\texttt{TimesFM}~\cite{das2023decoder} is a time series foundation model that adopts a decoder-style attention mechanism for pretraining. It processes time series data by segmenting it into patches, enabling \texttt{TimesFM} to capture both local and global temporal dependencies. 
\texttt{Timer}~\cite{liu2024timer} curates large-scale datasets with up to 1 billion time points, unifying heterogeneous time series into a single series into single-series sequence (S3) format, thus developing a GPT-style architecture for large time series models (LTSMs). 
\texttt{Moirai}~\cite{woo2024unified} extends the \texttt{TimesFM} architecture to the any-variate setting by ``flattening'' multivariate time series to a long single sequence. It achieves universal time series forecasting by training a masked encoder-based Transformer to address three challenges inherent in time series data: i) multiple frequencies, ii) multivariate forecasting, and iii) varying distributions. 
\texttt{Chronos}~\cite{ansari2024chronos} tokenizes time series values using scaling and quantization into a fixed vocabulary and trains the transformer-based architecture via a cross-entropy loss function.

\subsection{Forecasting Task}
Assuming we are given a dataset where each entry consists of data from $w$ past days (``window size'') including water levels, flows, and other measurements for multiple stations, our goal is to predict water levels for the following $k$ days (``lead time'') at stations of interest. We formulate the forecasting problem as:
\begin{equation}
\label{eq:problem_formulation}
    \mathcal{F}_{\theta}(\textbf{X}^{all}_{t-w+1:t}) \rightarrow  \textbf{Y}^{water}_{t+1:t+k},
\end{equation}
where $\mathcal{F}_{\theta}$ represents a predictive model with learnable parameters $\theta$. The superscripts $all$ and $water$ represent all (e.g., water levels, flows, and other measurements) and water variables, respectively.
In what follows, we will drop the superscripts and use $\textbf{X}_{t-w+1:t}$ and $\textbf{Y}_{t+1:t+k}$.

\paragraph{Input Features \& Output Targets.}
We task the model with an instance of supervised regression. 
To this end, we define the input feature matrix at time step $t$ as $\textbf{X}_{t-w+1:t}=[\textbf{X}_{t-w+1}, \dots, \textbf{X}_{t-1}, \textbf{X}_{t}] \in \mathcal{R}^{w \times n}$ and the output target as $\textbf{Y}_{t+1:t+k}=[\textbf{Y}_{t+1}, \textbf{Y}_{t+2}, \dots, \textbf{Y}_{t+k}] \in \mathcal{R}^{k \times m}$, where $n$ and $m$ denote the number of input and output variables. 
We collected all input-output pairs to form the dataset $\mathcal{D}=\{(\textbf{X}_{t-w+1:t}, \textbf{Y}_{t+1:t+k})\}^{T-k}_{t=w}$.

\begin{table*}[ht!]
\centering
    \resizebox{1.81\columnwidth}{!}{
    \begin{tabular}{c|c|cc|cc|cc|cc|cc|cc}
        \toprule
        \multirow{2}{*}{Models} & \multirow{2}{*}{\makecell[c]{Lead\\Time}} 
        & \multicolumn{2}{c|}{NP205} & \multicolumn{2}{c|}{P33} 
        & \multicolumn{2}{c|}{G620} & \multicolumn{2}{c|}{NESRS1} 
        & \multicolumn{2}{c|}{NESRS2} & \multicolumn{2}{c}{Overall} \\
        \cmidrule(lr){3-4} \cmidrule(lr){5-6} \cmidrule(lr){7-8} 
        \cmidrule(lr){9-10} \cmidrule(lr){11-12} \cmidrule(lr){13-14}
        & & MAE & RMSE & MAE & RMSE & MAE & RMSE & MAE & RMSE & MAE & RMSE & MAE & RMSE \\
                
        \midrule
        \midrule
        \multirow{4}{*}{\makecell[c]{NLinear\\ \cite{zeng2023transformers}\\Linear-based}} & 7  & 0.195 & 0.528 & 0.114 & 0.508 & 0.086 & 0.132 & 0.081 & 0.176 & 0.063 & 0.113 & 0.108 & 0.291 \\
                & 14 & 0.207 & 0.313 & 0.095 & 0.135 & 0.108 & 0.154 & 0.087 & 0.134 & 0.078 & 0.131 & 0.115 & 0.174 \\
                & 21 & 0.257 & 0.387 & 0.120 & 0.161 & 0.142 & 0.189 & 0.110 & 0.158 & 0.102 & 0.157 & 0.146 & 0.210 \\
                & 28 & 0.323 & 0.448 & 0.151 & 0.189 & 0.183 & 0.226 & 0.138 & 0.182 & 0.127 & 0.179 & 0.185 & 0.245 \\

        \midrule
        \multirow{4}{*}{\makecell[c]{DLinear\\ \cite{zeng2023transformers}\\Linear-based}} & 7  & 0.165 & 0.245 & 0.081 & 0.117 & 0.094 & 0.132 & 0.071 & 0.118 & 0.062 & 0.110 & 0.095 & 0.145 \\
                & 14 & 0.236 & 0.351 & 0.125 & 0.170 & 0.130 & 0.174 & 0.127 & 0.184 & 0.127 & 0.183 & 0.149 & 0.212 \\
                & 21 & 0.352 & 0.463 & 0.227 & 0.297 & 0.198 & 0.258 & 0.266 & 0.332 & 0.268 & 0.331 & 0.262 & 0.336 \\
                & 28 & 0.474 & 0.598 & 0.360 & 0.444 & 0.278 & 0.367 & 0.425 & 0.500 & 0.424 & 0.498 & 0.392 & 0.482 \\        
                
        \midrule
        \multirow{4}{*}{\makecell[c]{NBEATS\\ \cite{oreshkin2019n}\\MLP-based}} & 7  & 0.149 & 0.223 & 0.062 & 0.106 & 0.064 & 0.116 & 0.056 & 0.107 & 0.049 & 0.099 & 0.076 & 0.130 \\
                & 14 & 0.195 & 0.312 & 0.090 & 0.140 & 0.097 & 0.159 & 0.083 & 0.140 & 0.078 & 0.138 & 0.109 & 0.178 \\
                & 21 & 0.262 & 0.415 & 0.111 & 0.167 & 0.123 & 0.194 & 0.101 & 0.167 & 0.097 & 0.169 & 0.139 & 0.222 \\
                & 28 & 0.319 & 0.468 & 0.139 & 0.198 & 0.150 & 0.227 & 0.136 & 0.201 & 0.133 & 0.203 & 0.176 & 0.259 \\
                
        \midrule  
        \multirow{4}{*}{\makecell[c]{TimeMixer\\ \cite{wang2024timemixer}\\MLP-based}} & 7  & 0.170 & 0.290 & 0.079 & 0.121 & 0.070 & 0.126 & 0.069 & 0.119 & 0.061 & 0.113 & 0.090 & 0.154 \\
                & 14 & 0.249 & 0.358 & 0.125 & 0.168 & 0.160 & 0.200 & 0.109 & 0.160 & 0.110 & 0.163 & 0.151 & 0.210 \\
                & 21 & 0.482 & 0.692 & 0.195 & 0.252 & 0.248 & 0.327 & 0.187 & 0.239 & 0.167 & 0.219 & 0.256 & 0.346 \\
                & 28 & 0.622 & 0.807 & 0.228 & 0.287 & 0.269 & 0.350 & 0.232 & 0.287 & 0.208 & 0.266 & 0.312 & 0.399 \\

        \midrule
        \multirow{4}{*}{\makecell[c]{TSMixer\\ \cite{chen2023tsmixer}\\ MLP-based}} & 7  & 0.145 & 0.226 & 0.072 & 0.111 & 0.080 & 0.126 & 0.063 & 0.108 & 0.055 & 0.100 & 0.083 & 0.134 \\
                & 14 & 0.213 & 0.328 & 0.100 & 0.142 & 0.115 & 0.167 & 0.089 & 0.135 & 0.082 & 0.133 & 0.204 & 0.181 \\
                & 21 & 0.271 & 0.401 & 0.126 & 0.169 & 0.149 & 0.202 & 0.112 & 0.158 & 0.106 & 0.157 & 0.153 & 0.217 \\
                & 28 & 0.339 & 0.465 & 0.153 & 0.200 & 0.185 & 0.243 & 0.130 & 0.179 & 0.122 & 0.173 & 0.186 & 0.252 \\
                
        \midrule                                
        \multirow{4}{*}{\makecell[c]{TSMixerx\\ \cite{chen2023tsmixer}\\MLP-based}} & 7  & 0.366 & 0.570 & 0.154 & 0.205 & 0.171 & 0.234 & 0.139 & 0.211 & 0.129 & 0.204 & 0.192 & 0.285 \\
                & 14 & 0.331 & 0.405 & 0.163 & 0.203 & 0.205 & 0.268 & 0.167 & 0.230 & 0.157 & 0.222 & 0.204 & 0.266 \\
                & 21 & 0.702 & 1.138 & 0.242 & 0.350 & 0.270 & 0.398 & 0.236 & 0.335 & 0.222 & 0.301 & 0.334 & 0.504 \\
                & 28 & 0.560 & 0.841 & 0.293 & 0.409 & 0.350 & 0.492 & 0.312 & 0.431 & 0.275 & 0.368 & 0.358 & 0.508 \\

        \midrule
        \multirow{4}{*}{\makecell[c]{Informer\\ \cite{zhou2021informer}\\Transformer-based}}& 7  & 0.279 & 0.389 & 0.134 & 0.174 & 0.160 & 0.207 & 0.155 & 0.193 & 0.155 & 0.195 & 0.177 & 0.232 \\
                & 14 & 0.385 & 0.492 & 0.262 & 0.314 & 0.237 & 0.311 & 0.415 & 0.453 & 0.423 & 0.456 & 0.344 & 0.405 \\
                & 21 & 0.438 & 0.591 & 0.358 & 0.403 & 0.340 & 0.372 & 0.152 & 0.198 & 0.145 & 0.195 & 0.287 & 0.352 \\
                & 28 & 0.627 & 0.781 & 0.497 & 0.553 & 0.658 & 0.728 & 0.302 & 0.363 & 0.302 & 0.357 & 0.478 & 0.557 \\
                
        \midrule
        \multirow{4}{*}{\makecell[c]{PatchTST\\ \cite{nie2022time}\\Transformer-based}} & 7  &0.132 &0.208 & 0.067 &0.103 & 0.071 & 0.113 & 0.060 & 0.105 & 0.052 & 0.097 & 0.076 & 0.125 \\
                & 14 & 0.211 &  0.317 & 0.096 & 0.137 & 0.109 & 0.156 & 0.087 & 0.136 & 0.079 & 0.132 & 0.116 & 0.176 \\
                & 21 & 0.280 & 0.406 & 0.127 & 0.168 & 0.151 & 0.198 & 0.114 & 0.162 & 0.105 & 0.161 & 0.156 & 0.219 \\
                & 28 & 0.350 & 0.480 & 0.155 & 0.197 & 0.191 & 0.239 & 0.140 & 0.188 & 0.129 & 0.184 & 0.193 & 0.258 \\
                
        \midrule
        \multirow{4}{*}{\makecell[c]{iTransformer\\ \cite{liu2023itransformer}\\Transformer-based}} & 7  & 0.205 & 0.320 & 0.080 & 0.120 & 0.083 & 0.135 & 0.072 & 0.118 & 0.070 & 0.111 & 0.102 & 0.161 \\
                & 14 & 0.282 & 0.416 & 0.121 & 0.165 & 0.131 & 0.184 & 0.113 & 0.161 & 0.105 & 0.150 & 0.150 & 0.215 \\
                & 21 & 0.310 & 0.455 & 0.127 & 0.175 & 0.154 & 0.212 & 0.129 & 0.178 & 0.123 & 0.171 & 0.169 & 0.238 \\
                & 28 & 0.335 & 0.503 & 0.156 & 0.206 & 0.193 & 0.263 & 0.160 & 0.215 & 0.144 & 0.201 & 0.198 & 0.278 \\
                
        \midrule
        \multirow{4}{*}{\makecell[c]{KAN\\ \cite{liu2024kan}\\KAN-based Model}} & 7  & 0.176 & 0.268 & 0.085 & 0.131 & 0.101 & 0.146 & 0.076 & 0.130 & 0.071 & 0.123 & 0.102 & 0.160 \\
                & 14 & 0.257 & 0.374 & 0.126 & 0.169 & 0.148 & 0.203 & 0.116 & 0.168 & 0.110 & 0.165 & 0.151 & 0.216 \\
                & 21 & 0.292 & 0.441 & 0.144 & 0.196 & 0.160 & 0.225 & 0.138 & 0.200 & 0.133 & 0.199 & 0.174 & 0.252 \\
                & 28 & 0.357 & 0.502 & 0.165 & 0.216 & 0.188 & 0.258 & 0.185 & 0.243 & 0.178 & 0.240 & 0.214 & 0.292 \\
                
        \midrule
        \multirow{4}{*}{\makecell[c]{RMoK\\ \cite{han2024kan4tsf}\\KAN-based Model}} & 7  & 0.171 & 0.247 & 0.088 & 0.118 & 0.100 & 0.137 & 0.080 & 0.116 & 0.070 & 0.106 & 0.102 & 0.145 \\
                & 14 & 0.264 & 0.365 & 0.129 & 0.160 & 0.152 & 0.190 & 0.110 & 0.148 & 0.099 & 0.139 & 0.151 & 0.200 \\
                & 21 & 0.310 & 0.438 & 0.145 & 0.184 & 0.179 & 0.226 & 0.133 & 0.172 & 0.124 & 0.170 & 0.178 & 0.238 \\
                & 28 & 0.337 & 0.483 & 0.153 & 0.202 & 0.185 & 0.250 & 0.143 & 0.194 & 0.134 & 0.185 & 0.191 & 0.263 \\

        \midrule
        \multirow{4}{*}{\makecell[c]{TimeLLM\\ \cite{jin2023time}\\LLM-based Model}} & 7  & 0.228 & 0.308 & 0.106 & 0.135 & 0.125 & 0.160 & 0.090 & 0.124 & 0.080 & 0.121 & 0.126 & 0.170 \\
                & 14 & 0.311 & 0.440 & 0.143 & 0.177 & 0.185 & 0.221 & 0.130 & 0.171 & 0.118 & 0.169 & 0.178 & 0.236 \\
                & 21 & 0.349 & 0.489 & 0.166 & 0.200 & 0.211 & 0.248 & 0.146 & 0.189 & 0.132 & 0.184 & 0.201 & 0.262 \\
                & 28 & 0.413 & 0.542 & 0.196 & 0.232 & 0.258 & 0.298 & 0.179 & 0.217 & 0.162 & 0.206 & 0.242 & 0.299 \\

        \midrule
        \midrule
        \multirow{4}{*}{\makecell[c]{TimeGPT\\ \cite{garza2023timegpt}\\Foundational Model}} & 7  & 0.160 & 0.262 & 0.082 & 0.138 & 0.086 & 0.153 & 0.074 & 0.136 & 0.065 & 0.131 & 0.093 & 0.164 \\
                & 14 & 0.247 & 0.400 & 0.111 & 0.165 & 0.120 & 0.181 & 0.102 & 0.167 & 0.090 & 0.165 & 0.134 & 0.216 \\
                & 21 & 0.298 & 0.492 & 0.143 & 0.217 & 0.159 & 0.246 & 0.132 & 0.215 & 0.126 & 0.222 & 0.172 & 0.278 \\
                & 28 & 0.410 & 0.608 & 0.204 & 0.263 & 0.229 & 0.310 & 0.179 & 0.259 & 0.166 & 0.273 & 0.238 & 0.342 \\
                
        \midrule
        \multirow{4}{*}{\makecell[c]{TimesFM\\ \cite{das2023decoder}\\Foundational Model}} & 7  & 0.224 & 0.337 & 0.118 & 0.166 & 0.121 & 0.186 & 0.108 & 0.171 & 0.094 & 0.156 & 0.133 & 0.203 \\
                & 14 &  0.376 & 0.584 & 0.182 & 0.244 & 0.177 & 0.269 & 0.153 & 0.229 & 0.154 & 0.230 & 0.208 & 0.311 \\
                & 21 & 0.536 & 0.759 & 0.246 & 0.310 & 0.256 & 0.342 & 0.211 & 0.282 & 0.206 & 0.282 & 0.291 & 0.395 \\
                & 28 & 0.639 & 0.869 & 0.291 & 0.350 & 0.299 & 0.381 & 0.246 & 0.316 & 0.234 & 0.301 & 0.342 & 0.443 \\

        \midrule
        \multirow{4}{*}{\makecell[c]{Timer\\ \cite{liu2024timer}\\Foundational Model}} & 7  & 0.293 & 0.374 & 0.160 & 0.194 & 0.181 & 0.224 & 0.146 & 0.184 & 0.131 & 0.171 & 0.182 & 0.229 \\
                & 14 & 0.456 & 0.594 & 0.238 & 0.277 & 0.291 & 0.344 & 0.212 & 0.251 & 0.193 & 0.235 & 0.278 & 0.340 \\
                & 21 & 0.559 & 0.720 & 0.293 & 0.330 & 0.386 & 0.433 & 0.265 & 0.299 & 0.230 & 0.270 & 0.347 & 0.410 \\
                & 28 & 0.611 & 0.784 & 0.322 & 0.366 & 0.446 & 0.501 & 0.294 & 0.331 & 0.251 & 0.291 & 0.385 & 0.455 \\
                
        \midrule
        \multirow{4}{*}{\makecell[c]{Moirai\\ \cite{woo2024unified}\\Foundational Model}} & 7  & 0.471 & 0.742 & 0.205 & 0.299 & 0.285 & 0.424 & 0.186 & 0.184 & 0.161 & 0.231 & 0.261 & 0.389 \\
                & 14 & 0.483 & 0.755 & 0.212 & 0.305 & 0.294 & 0.429 & 0.187 & 0.264 & 0.165 & 0.242 & 0.265 & 0.397 \\
                & 21 & 0.551 & 0.817 & 0.328 & 0.418 & 0.432 & 0.557 & 0.255 & 0.343 & 0.223 & 0.309 & 0.358 & 0.489 \\
                & 28 & 0.555 & 0.820 & 0.331 & 0.422 & 0.446 & 0.569 & 0.261 & 0.351 & 0.224 & 0.305 & 0.364 & 0.493 \\
                
        \midrule   
        \multirow{4}{*}{\makecell[c]{\textbf{Chronos}\\ \cite{ansari2024chronos} \\Foundational Model }} & 7  & \textbf{0.094} & \textbf{0.132} & \textbf{0.047} & \textbf{0.059} & \textbf{0.042} & \textbf{0.054} & \textbf{0.036} & \textbf{0.051} & \textbf{0.028} & \textbf{0.045} & \textbf{0.049} & \textbf{0.075} \\
                & 14 & \textbf{0.127} & \textbf{0.176} & \textbf{0.069} & \textbf{0.089} & \textbf{0.060} & \textbf{0.073} & \textbf{0.053} & \textbf{0.073} & \textbf{0.037} & \textbf{0.058} &\textbf{0.069} & \textbf{0.103} \\
                & 21 & \textbf{0.148} & \textbf{0.198} & \textbf{0.087} & \textbf{0.107} & \textbf{0.074} & \textbf{0.089} & \textbf{0.069} & \textbf{0.089} & \textbf{0.047} & \textbf{0.066} &\textbf{0.085} & \textbf{0.119} \\
                & 28 & \textbf{0.147} & \textbf{0.200} & \textbf{0.090} & \textbf{0.116} & \textbf{0.083} & \textbf{0.097} & \textbf{0.071} & \textbf{0.095} & \textbf{0.049} & \textbf{0.068} &\textbf{0.088} & \textbf{0.124} \\
        \midrule     
        \bottomrule
    \end{tabular}
}
    \caption{Performance across 5 stations (NP205, P33, G620, NESRS1, NESRS2) for lead times of 7, 14, 21, and 28 days. The first 12 models are task-specific, while the last 5 are pre-trained foundation models for time series. The best results are in \textbf{bold}. }
    \label{tab:overall_perform}
\end{table*}

\section{Experiments}

\subsection{Experimental Setup}

\paragraph{Dataset splitting.} The data was split into training, validation, and test sets with an approximate ratio of $70\%:15\%:15\%$. The training set includes data from October 16, 2020, to July 1, 2023, representing 989 days. The validation set spans July 2, 2023, to January 28, 2024, covering 211 days, and the test set includes data from January 29, 2024, to August 26, 2024, also comprising 211 days.

\paragraph{Task Settings.}
To ensure consistent evaluation across different models, we considered all variables from the previous 100 days as input and trained all task-specific models to forecast water levels at 5 stations (NP205, P33, G620, NESRS1, and NESRS2, as shown in Figure \ref{fig:everglade_map}) for the next 7, 14, 21, and 28 days. Time Series Foundation Models are with the same setting in zero-shot inference.

\paragraph{Evaluation Metrics.}
Following the work \cite{shi2023deep,shi2025fidlar}, we used Mean Absolute Error (MAE) and Root Mean Squared Error (RMSE) to evaluate the overall performance of these models. 
MAE measures the predictive robustness of an algorithm but is relatively insensitive to outliers, whereas RMSE can amplify errors.
Additionally, we also include the Symmetric Extremal Dependence Index (SEDI) as a specialized metric for extreme values \cite{han2024far,zheng2025sf}.
It classifies each prediction as either ``extreme'' or ``normal'' based on upper or lower quantile thresholds. 
SEDI values belong to the range $[0,1]$, where higher values indicate better accuracy in identifying extreme water levels.

\begin{equation}
\text{MAE} = \frac{1}{N} \sum_{i=1}^N \left| y_i - \hat{y}_i \right|, \text{and}
\end{equation}
\begin{equation}
\text{RMSE} = \sqrt{\frac{1}{N} \sum_{i=1}^N \left( y_i - \hat{y}_i \right)^2}, \text{and}
\end{equation}
\begin{equation}
\text{SEDI}(p) = \frac{\Sigma(\hat{y} < y_{\text{low}}^p \& y < y_{\text{low}}^p) + \Sigma(\hat{y} > y_{\text{up}}^p \& y > y_{\text{up}}^p)}{\Sigma(y < y_{\text{low}}^p) + \Sigma(y > y_{\text{up}}^p)},
\end{equation}
where $y$ and $\hat{y}$ denote the ground truth and prediction, respectively. The tests, $\hat{y} < y_{\text{up}}^p$ and $y < y_{\text{up}}^p$ judge whether the predicted and observed values are extremes based on the threshold value $y_{\text{up}}^p$. Following the work \cite{han2024weather}, we set $10\%$ and $90\%$ as the low and high thresholds, respectively. $N$ is the number of samples.

\paragraph{Implementation Details.}
For task-specific models, we first pre-train them before conducting inference.
The training is performed for a total of 1,000 epochs, starting with a learning rate of $1e^{-3}$. We used a batch size of 32 for training, and early stopping is executed if training loss does not decrease for 50 iterations.
For pre-trained time series foundation models, we directly used their existing model weights for zero-shot inference on our test set.
Experiments were conducted on an NVIDIA A100 GPU with 80GB memory.
More details are provided in Appendix \ref{sec:imple_detail}.

\section{Results}
Table \ref{tab:overall_perform} compares the quantitative performance of 12 task-specific and 5 foundation models across 5 stations and over 4 lead times, allowing us to address the following questions.

\paragraph{RQ1: How do time series foundation models perform compared to task-specific models?}
Surprisingly, the time series foundation model, \texttt{Chronos}, consistently outperforms other models over various lead times across all stations, showcasing exceptional performance in zero-shot inference.
Temporally, while the performance of all models declines as the lead time increases, \texttt{Chronos} demonstrates a comparatively slower drop.
Spatially, we observe the performance of all models varies across stations with the worst scenarios for the NP205 station, where \texttt{Chronos} still achieves the lowest prediction errors.
A possible reason for the lower performance at the NP205 station is its weaker correlation with other water stations (see Figure \ref{fig:water_correlation} in Appendix \ref{sec:Correlation}).
Note that the remaining foundation models show relatively poor performance. \texttt{TimeGPT} achieves results comparable to the best task-specific models, \texttt{NBEATS} and \texttt{PatchTST}, while \texttt{TimesFM}, \texttt{Timer} and \texttt{Moirai} perform much worse.
Due to the limited space, we provide the visualization and qualitative analysis in Appendix \ref{sec:more_visual}.

\paragraph{RQ2: How do the task-specific models perform?}
The performance varies across their model architectures, with \texttt{NBEATS}, \texttt{TSMixer}, \texttt{PatchTST}, and \texttt{RMoK} consistently outperforming others \emph{within} their respective categories.
\texttt{RMoK} model outperforms the vanilla \texttt{KAN}, particularly for long-term forecasting, highlighting the effectiveness of the mixture-of-experts approach.
Furthermore, the linear-based models excel in short-term predictions but experience a significant drop in performance as the forecasting horizon increases. In contrast, MLP-based \texttt{NBEATS} and Transformer-based \texttt{PatchTST} help mitigate this issue.

\paragraph{RQ3: How do the 17 models perform with respect to extreme values?}
Following the work for extreme weather prediction~\cite{han2024weather}, we use the SEDI metric to evaluate model performance under extreme values and report the results in Table \ref{tab:extreme_perform}. 
We observed that \texttt{Chronos} significantly outperforms other models.
Moreover, even though \texttt{TSMixerx} shows average performance in overall water level prediction (see Table \ref{tab:overall_perform}), it performs notably well for predicting extreme water levels at the P33, G620, and NESRS1 stations. 
To better understand this observation, we visualize the ground truth and model predictions with a lead time of 28 days (see Figure \ref{fig:three_stations} in Appendix \ref{sec:three_stations}). 
The visualizations reveal that \texttt{TSMixerx} generally follows the global trend but fails to capture the temporal details (e.g., abrupt changes around time points 50, 130, 150).
Notably, despite producing several false alarms between time 140 and 170, \texttt{TSMixerx} successfully predicts most extreme values highlighted in red circles. 
\begin{table}[ht!]
\centering
\resizebox{0.99\columnwidth}{!}{%
\begin{tabular}{l|ccccc|c}
\toprule
Models           & NP205 & P33 & G620 & NESRS1 & NESRS2 & Overall \\
\midrule
NLinear         & 0.143     & 0.122   & 0.146    & 0.122 & 0.167 & 0.140\\
DLinear         & 0.262     & 0.512   & 0.390    & 0.488 & 0.500 & 0.430\\
NBEATS          & 0.262     & 0.268   & 0.463   & 0.268 & 0.286  & 0.310\\
TimeMixer       & 0.119     & 0.049   & 0.341    & 0.000 & 0.000 & 0.102\\
TSMixer         & 0.143     & 0.146   & 0.195    & 0.073 & 0.095 & 0.131\\
TSMixerx        & 0.381     & \textbf{0.829}   & \textbf{0.829}    & \textbf{0.780} & 0.476 & 0.659 \\
Informer        & 0.500     & 0.488   & 0.487    & 0.512 & 0.500  & 0.498\\
PatchTST        & 0.190     & 0.146   & 0.195    & 0.146 & 0.190  & 0.174\\
iTransformer    & 0.143     & 0.244   & 0.244    & 0.244 & 0.595  & 0.294\\
KAN             & 0.119     & 0.073   & 0.341    & 0.171 & 0.190  & 0.179\\
RMoK            & 0.119     & 0.098   & 0.171    & 0.317 & 0.309  & 0.203\\
TimeLLM         & 0.001     & 0.001   & 0.001    & 0.001 & 0.001  & 0.001\\
\midrule
TimeGPT         & 0.024     & 0.195   & 0.195    & 0.171 & 0.167  & 0.150\\
TimesFM         & 0.095     & 0.024   & 0.170    & 0.000 & 0.000  & 0.058\\
Timer         & 0.000     & 0.000   & 0.000    & 0.000 & 0.000  & 0.000\\
Moirai           & 0.476     & 0.317   & 0.488    & 0.122 & 0.023  & 0.285\\
\textbf{Chronos}& \textbf{0.786}     & 0.659   & 0.805    & 0.537 & \textbf{0.762} & \textbf{0.710}\\
\bottomrule
\end{tabular}%
}
\caption{SEDI values (the higher, the better) for predicting extreme values across 5 stations with lead time of 28 days. }
\label{tab:extreme_perform}
\end{table}

\begin{figure}[htbp]
\centering
    \includegraphics[width=\columnwidth]{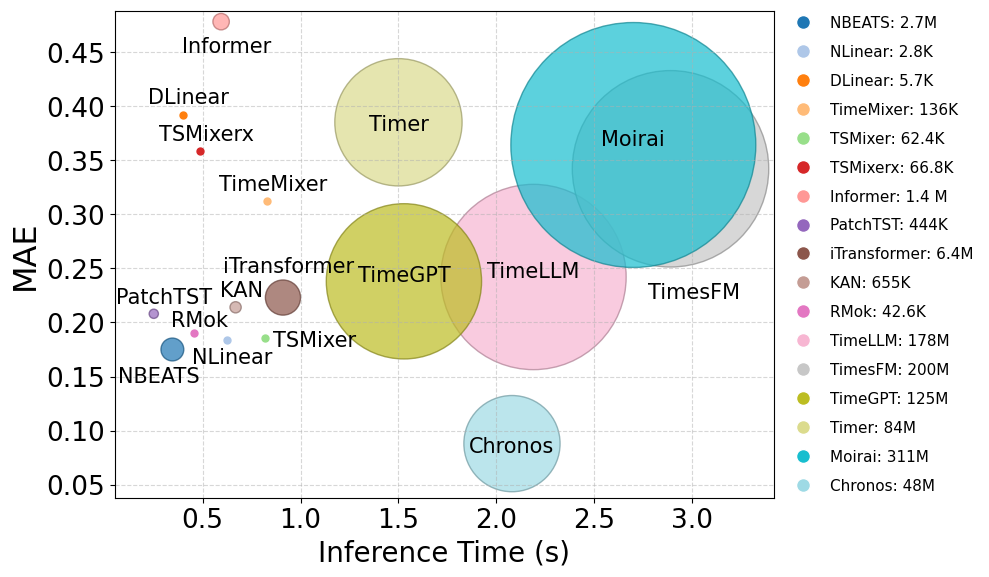}
\caption{Plot of Accuracy vs Efficiency vs Model Size for an input length of 100 days and lead time of 28 days.}
\vspace{-5mm}
\label{fig:ModelComplexity}
\end{figure}

\paragraph{RQ4: How do models perform with different model sizes?}
Figure~\ref{fig:ModelComplexity} compares models by visualizing their prediction accuracy (MAE), inference time, and model complexity. 
We found that while the model size influences predictive accuracy and inference time, it does not solely determine overall performance. 
For task-specific models, despite having a smaller size, \texttt{NBEATS} outperforms \texttt{Informer} and \texttt{TimesLLM}, both of which are larger. 
A similar phenomenon is observed between \texttt{Chronos} and other foundation models.
However, when comparing task-specific models to foundation models, we found that only \texttt{Chronos} significantly surpasses task-specific models in performance. 
This suggests that simply scaling model parameters may not improve prediction accuracy, guiding researchers to explore alternative strategies to optimize foundation models, such as expanding datasets covering more domains and designing more efficient network architectures. 

\paragraph{RQ5: What makes \texttt{Chronos} the highest performing model?}
While one could attribute the superior performance of \texttt{Chronos} to its \emph{tokenization} via scaling and quantization or the use of \emph{cross entropy} in training, a simpler explanation could be that it was exposed to training data that correlates with data from the Everglades National Park.
Even though there are no actual Everglades data included, we found that multiple weather datasets were used for its pre-training~\cite{ansari2024chronos}.
One of them contains daily measurements of precipitation, snow, snow depth, and temperature from climate stations located in 48 states in the USA. 
However, the other four foundation models in this study were not pre-trained on this dataset or similar datasets.

\begin{figure}[ht!]
\centering
    \includegraphics[width=0.98\columnwidth]{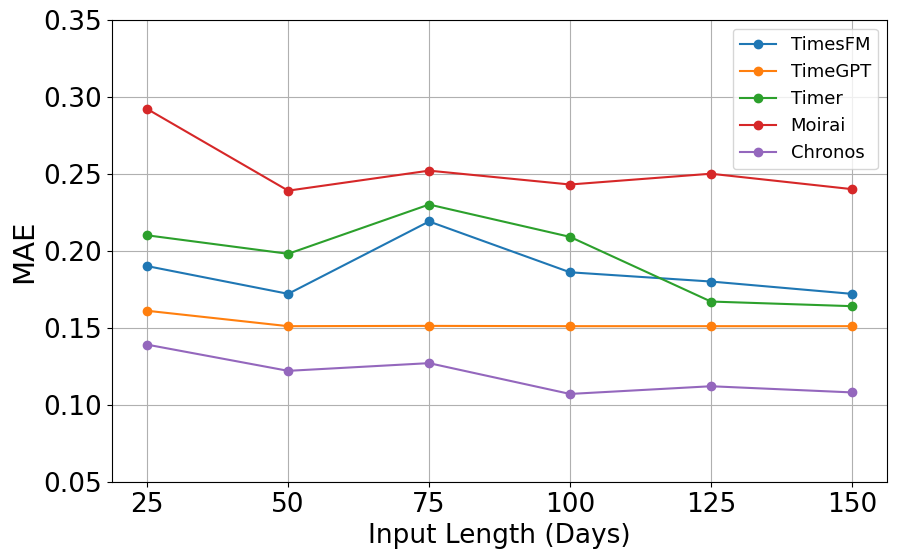}
\caption{Performance (lead time: 28 days) vs Input Length. We computed MAEs of ten samples for each input length.}
\label{fig:Parameter-Study}
\end{figure}

\paragraph{RQ6: Does input length impact the predictive accuracy of the models?}
One key advantage of foundation models is their ability to generate predictions for inputs of varying lengths without being retrained. 
To analyze the impact of input length on prediction performance, we performed experiments predicting 28 days ahead with varying input lengths and reported the results in Figure \ref{fig:Parameter-Study}. 
It shows that the prediction accuracy depends on the input lengths. The MAE values drop as the input length increases from 25 to 100 days, after which the performance stabilizes. 

\section{Related Work}
\paragraph{Physics-based Models.}
Physics-based methods have long been a cornerstone in hydrologic and hydraulics research. For example, 1D and 2D HEC-RAS models have been used to predict water levels \cite{munoz2022inter} and to simulate sewer and street flows in urban systems \cite{huang2023alternative}.
The South Florida Water Management Model (SFWMM) serves as a regional hydrological model designed to simulate water levels in the Everglades \cite{sfwmd2005documentation}.
Furthermore, researchers have also combined physics-based approaches with ML techniques, such as integrating a hydrologic-hydrodynamic coupling (H2C) model with 
long short-term memory (LSTM) networks to achieve highly accurate water level predictions in tidal reaches \cite{fei2023accurate}.
Despite their utility, these mechanistic models require complex variable integration and rely on predefined assumptions.

\paragraph{Statistical Models.} 
The Everglades Forecasting application (EverForecast) \cite{pearlstine2020near} is a statistical model designed to generate spatially continuous simulations of the Everglades water levels in South Florida.
Autoregressive Integrated Moving Average (ARIMA) model has been used for water level and discharge predictions \cite{agaj2024using}, while the Seasonal Autoregressive Integrated Moving Average (SARIMA) model \cite{azad2022water} was explored for forecasting Malaysia's mean sea level at coastal stations. 
The study \cite{sekban2022istanbul} focuses on predicting monthly dam water levels by comparing various statistical time series models. These models excel at identifying simple trends from historical data but are limited in capturing complex dynamics among multivariate time series.

\paragraph{Data-driven Machine Learning Models.} 
ML models such as support vector machines (SVM) \cite{khan2006application}, and extreme gradient boosting (XGBoost) \cite{nguyen2021development}  have demonstrated the ability to model nonlinear relationships in water level datasets.
We have also experimented with these two ML models, but the results were relatively poor.
Besides, the advent of deep learning has revolutionized hydrological forecasting, learning complex spatiotemporal patterns from large datasets. 
Artificial neural networks (ANNs) \cite{abiy2022multilayer}, convolutional neural networks (CNNs) \cite{bassah2025forecasting}, recurrent neural networks (RNNs) \cite{shi2023explainable}, Transformer models \cite{hou2024hydroformer}, and many hybrid models have been used to capture temporal dependencies in hydrological data. 
Furthermore, Transformer-based models, such as \texttt{Informer} \cite{zhou2021informer} and \texttt{PatchTST} \cite{nie2022time}, are proposed for time series modeling across various domains.
\texttt{NLinear} and \texttt{DLinear} are two models with only linear layers \cite{zeng2023transformers}. 
Neural Basis Expansion Analysis for Interpretable Time Series Forecasting (NBEATS) \cite{oreshkin2019n} and \texttt{TimeMixer} \cite{chen2023tsmixer} are two MLP-based architectures for multi-scale time series modeling.
Leveraging large language models (LLMs) for time series forecasting was investigated in \cite{jin2024position}, where a time series was treated as a sequence of tokens (values) and pre-trained LLMs were fine-tuned with well-designed prompts \cite{jin2023time}.

\paragraph{Time Series Foundation Models.}
\texttt{TimeGPT} \cite{garza2023timegpt}, a pioneering foundation model, employs a transformer-based architecture and was pre-trained on massive time series datasets, supporting both fine-tuning and zero-shot inference. \texttt{TimesFM} \cite{das2023decoder} follows a similar approach, utilizing a decoder-based architecture to improve time series forecasting. 
\texttt{Moment} \cite{goswami2024moment} introduces a multi-scale learning mechanism, enhancing its ability to model heterogeneous temporal dynamics. \texttt{Timer} \cite{liu2024timer} leverages a hybrid structure incorporating state-space models to improve long-term forecasting stability. \texttt{Moirai} \cite{woo2024unified} unifies multiple time series forecasting paradigms within a single foundation model framework. Finally, \texttt{Chronos} \cite{ansari2024chronos} introduces an innovative design that is optimized for zero-shot forecasting across numerous time series applications.
\section{Discussion}
\label{sec:discuss}
We evaluated 12 task-specific DL models and 5 pre-trained foundation models for a hydrological application, i.e., water level prediction in the Everglades. We summarized several takeaways. First,
the performance of task-specific models varies based on architecture and training. They require careful retraining whenever datasets or experimental settings change.
Second, not all of those foundation models demonstrate strong performance. \texttt{Chronos} with the zero-shot inference achieves the best performance, while others perform poorly. In addition to model architectures, a potential explanation is that \texttt{Chronos} has included training data that highly correlates with data from the Everglades, while the training data sets for other foundation models have a low correlation with data from the Everglades. 
Third, all models struggle to predict extremely high and low values accurately, suggesting that they may have difficulty capturing abrupt changes or sudden shifts in the underlying dynamics, potentially due to insufficient training samples for extreme events or inherent model biases. 
Future improvements could involve incorporating additional physical constraints \cite{yin2023physic,yin2025physics}, leveraging ensemble methods \cite{wu2023exploring}, or integrating uncertainty quantification \cite{shi2024codicast} to enhance robustness in extreme conditions.
Lastly, model performance varies across various locations, demanding context-aware, region-specific adaption to improve prediction performance \cite{li2024can}. 

\section{Conclusions}
\label{sec:conc}
Our findings highlight the potential applications of time series foundation models in hydrology.
The strong performance of \texttt{Chronos} suggests that certain pre-trained models can leverage their learned representations effectively, even in specialized applications.
Moreover, the flexibility of zero-shot inference across different input lengths underscores the high adaptability of foundation models, making them attractive for real-world deployment where data availability and sequence lengths vary.
However, further studies are needed to fully harness the potential of these foundation models for hydrologic applications. 
Exploring strategies to bridge the gap between general-purpose pre-training and the specific requirements of environmental forecasting presents a promising direction for future research.

\section*{Collaboration and Broader Impact}
ML and Everglades researchers were involved in the multidisciplinary collaborations. Domain experts helped in area selection, data acquisition, data preprocessing, and analysis of the results.
This work also helps domain researchers select the best AI models for water level forecasting in the Everglades and provides practical insights for government agencies interested in AI-driven hydrological applications, e.g., flood prediction and management.

\newpage
\bibliography{aaai2026}

\newpage
\section*{Reproducibility Checklist}
This paper:
\begin{itemize}
    \item Includes a conceptual outline and/or pseudocode description of AI methods introduced (\yes)
    \item Clearly delineates statements that are opinions, hypothesis, and speculation from objective facts and results (\yes)
    \item Provides well marked pedagogical references for less-familiare readers to gain background necessary to replicate the paper (\yes)
\end{itemize}

\noindent Does this paper make theoretical contributions? (\no)

\noindent If yes, please complete the list below. (answered \no)
\begin{itemize}
    \item All assumptions and restrictions are stated clearly and formally. (yes/partial/no)
    \item All novel claims are stated formally (e.g., in theorem statements). (yes/partial/no)
    \item Proofs of all novel claims are included. (yes/partial/no)
    \item Proof sketches or intuitions are given for complex and/or novel results. (yes/partial/no)
    \item Appropriate citations to theoretical tools used are given. (yes/partial/no)
    \item All theoretical claims are demonstrated empirically to hold. (yes/partial/no/NA)
    \item All experimental code used to eliminate or disprove claims is included. (yes/no/NA)
\end{itemize}

\noindent Does this paper rely on one or more datasets? (\yes)

\noindent If yes, please complete the list below.
\begin{itemize}
    \item A motivation is given for why the experiments are conducted on the selected datasets (\yes)
    \item All novel datasets introduced in this paper are included in a data appendix. (\yes)
    \item All novel datasets introduced in this paper will be made publicly available upon publication of the paper with a license that allows free usage for research purposes. (\yes)
    \item All datasets drawn from the existing literature (potentially including authors’ own previously published work) are accompanied by appropriate citations. (\yes)
    \item All datasets drawn from the existing literature (potentially including authors’ own previously published work) are publicly available. (\yes)
    \item All datasets that are not publicly available are described in detail, with explanation why publicly available alternatives are not scientifically satisficing. (\yes)
\end{itemize}

\noindent Does this paper include computational experiments? (\yes)

\noindent If yes, please complete the list below.
\begin{itemize}
    \item This paper states the number and range of values tried per (hyper-) parameter during development of the paper, along with the criterion used for selecting the final parameter setting. (\yes)
    \item Any code required for pre-processing data is included in the appendix. (\yes).
    \item All source code required for conducting and analyzing the experiments is included in a code appendix. (\yes)
    \item All source code required for conducting and analyzing the experiments will be made publicly available upon publication of the paper with a license that allows free usage for research purposes. (\yes)
    \item All source code implementing new methods have comments detailing the implementation, with references to the paper where each step comes from (\yes)
    \item If an algorithm depends on randomness, then the method used for setting seeds is described in a way sufficient to allow replication of results. (\yes)
    \item This paper specifies the computing infrastructure used for running experiments (hardware and software), including GPU/CPU models; amount of memory; operating system; names and versions of relevant software libraries and frameworks. (\yes)
    \item This paper formally describes evaluation metrics used and explains the motivation for choosing these metrics. (\yes)
    \item This paper states the number of algorithm runs used to compute each reported result. (\yes)
    \item Analysis of experiments goes beyond single-dimensional summaries of performance (e.g., average; median) to include measures of variation, confidence, or other distributional information. (\yes)
    \item The significance of any improvement or decrease in performance is judged using appropriate statistical tests (e.g., Wilcoxon signed-rank). (\yes)
    \item This paper lists all final (hyper-)parameters used for each model/algorithm in the paper’s experiments. (\yes)
\end{itemize}

\newpage
\appendix
\onecolumn
\section*{Appendix} 


\section{Data Set}
\label{sec:dataset}

\subsection{Data summary}
Table \ref{tab:data_summary} provides a summary of the key features included in the dataset used for the study. The data encompasses multiple hydrological and meteorological variables collected across various measuring stations. 
Daily rainfall data, measured in inches, is available for two measuring stations, namely NP205 and P33.
Potential Evapotranspiration (PET), measured in millimeters, represents the amount of water that would evaporate and transpire under normal conditions. PET is critical for understanding the water balance in the region as it helps to estimate the loss of water due to evaporation and transpiration. 
Daily pump and gate flow data is measured in cubic feet per second (cfs). 
Pumps and gates are used to regulate the water flow between different parts of the Everglades system, making the gate flow data essential for modeling water management operations. 
Daily water level data, recorded in feet is available for fourteen stations. 

\begin{table*}[ht]
    \centering
    \resizebox{\textwidth}{!}{ 
    \begin{tabular}{l|c|c|c|l}
        \toprule
        \textbf{Feature} & \textbf{Interval} & \textbf{Unit} & \textbf{\#Var.} & \textbf{Location} \\
        \midrule
        Rainfall & Daily & $inches$ & 2 & NP-205, P33 \\
        Potential Evapotranspiration & Daily & $mm$ & 2 & NP-205, P33 \\
        Pump flows & Daily & $cfs$ & 8 & S356, S332B, S332BN, S332C, S332DX1, S332D, S200, S199 \\
        Gate flows & Daily & $cfs$ & 11 & S343A, S343B, S12A, S12B, S12C, S12D, S333N, S333, S334, S344, S18C \\
        Water levels & Daily & $ft$ & 14 & S12A, S12B, S12C, S12D, S333, S334, NESRS1, NESRS2, NP-205, P33, G620, EVER4, TSH, NP62 \\
        \bottomrule
    \end{tabular}
    }
    \caption{Summary of the dataset.}
    \vspace{-3mm}
    \label{tab:data_summary}
\end{table*}

\subsection{Pre-processing}
\textbf{Data Cleaning.}
The raw water level data was reported in two different vertical datums: the North American Vertical Datum of 1988 (NAVD88) and the National Geodetic Vertical Datum of 1929 (NGVD29). 
Since direct comparisons require a consistent datum, adjustments were made following domain expert recommendations to standardize all values to NGVD29. 
For example, the G620 station requires an adjustment of +1.51 ft, while S333T requires +1.54 ft, among others.
These values were carefully selected based on the suggestions from domain experts.
\textbf{Temporal Alignment.} The data pre-processing ensured temporal consistency across all stations. Daily data points were synchronized to a uniform timestamp. 
%
\textbf{Data Interpolation.}  Completeness of data is vital for accurate forecasting of water levels, particularly in ecologically sensitive regions such as the Everglades National Park (ENP). 
There are certain gates and stations that still exhibit missing values. We employed time-based interpolation and backward fill techniques to address the missing data.

\section{Extreme Value Predictions}
\label{sec:three_stations}
\begin{figure}[ht!]
    \centering
    \vspace{-2mm}
    \begin{subfigure}[b]{0.33\textwidth}
        \includegraphics[width=\textwidth]{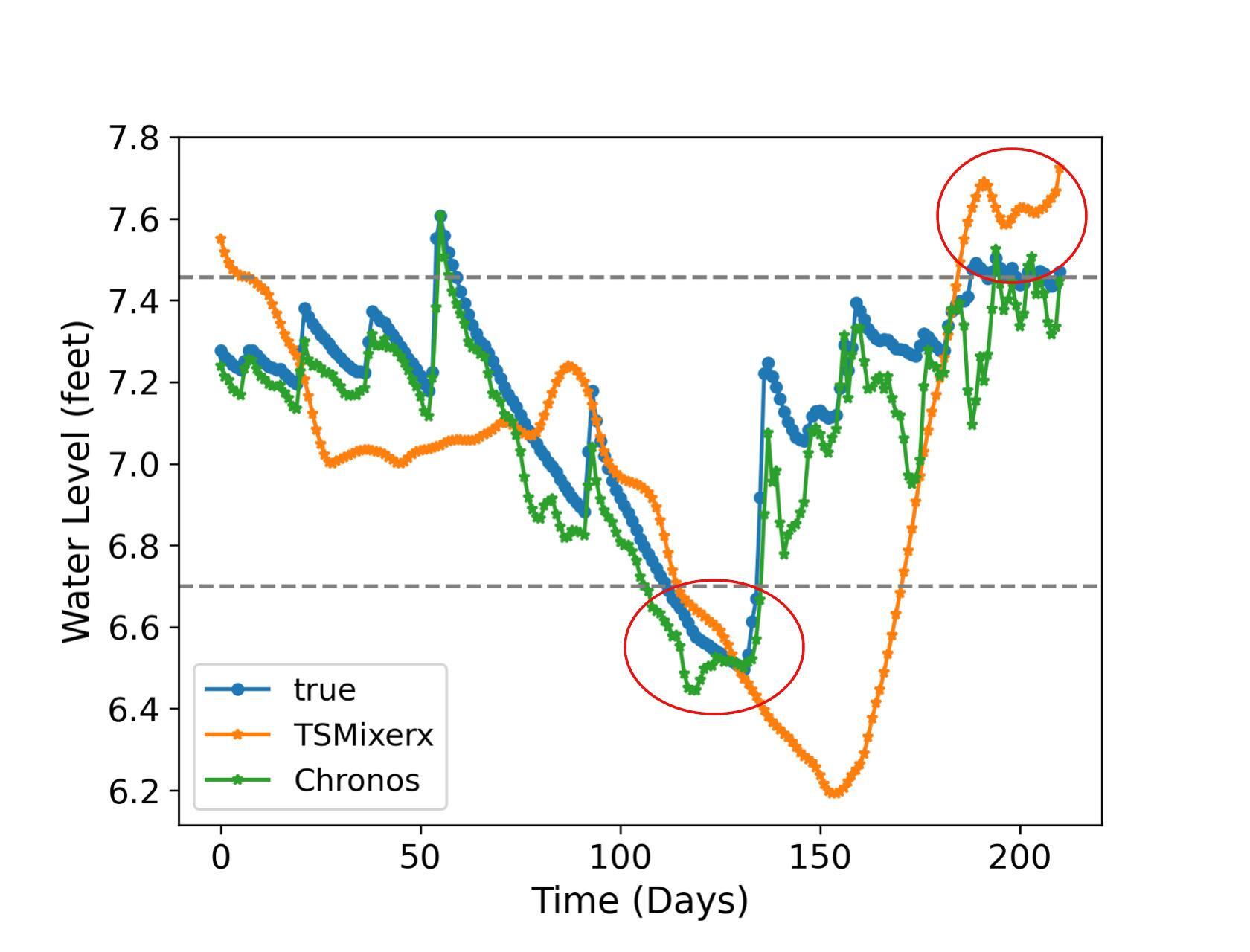}
        \caption{ }
        \label{fig:P33_visual}
    \end{subfigure}
    \hfill
    \begin{subfigure}[b]{0.33\textwidth}
        \includegraphics[width=\textwidth]{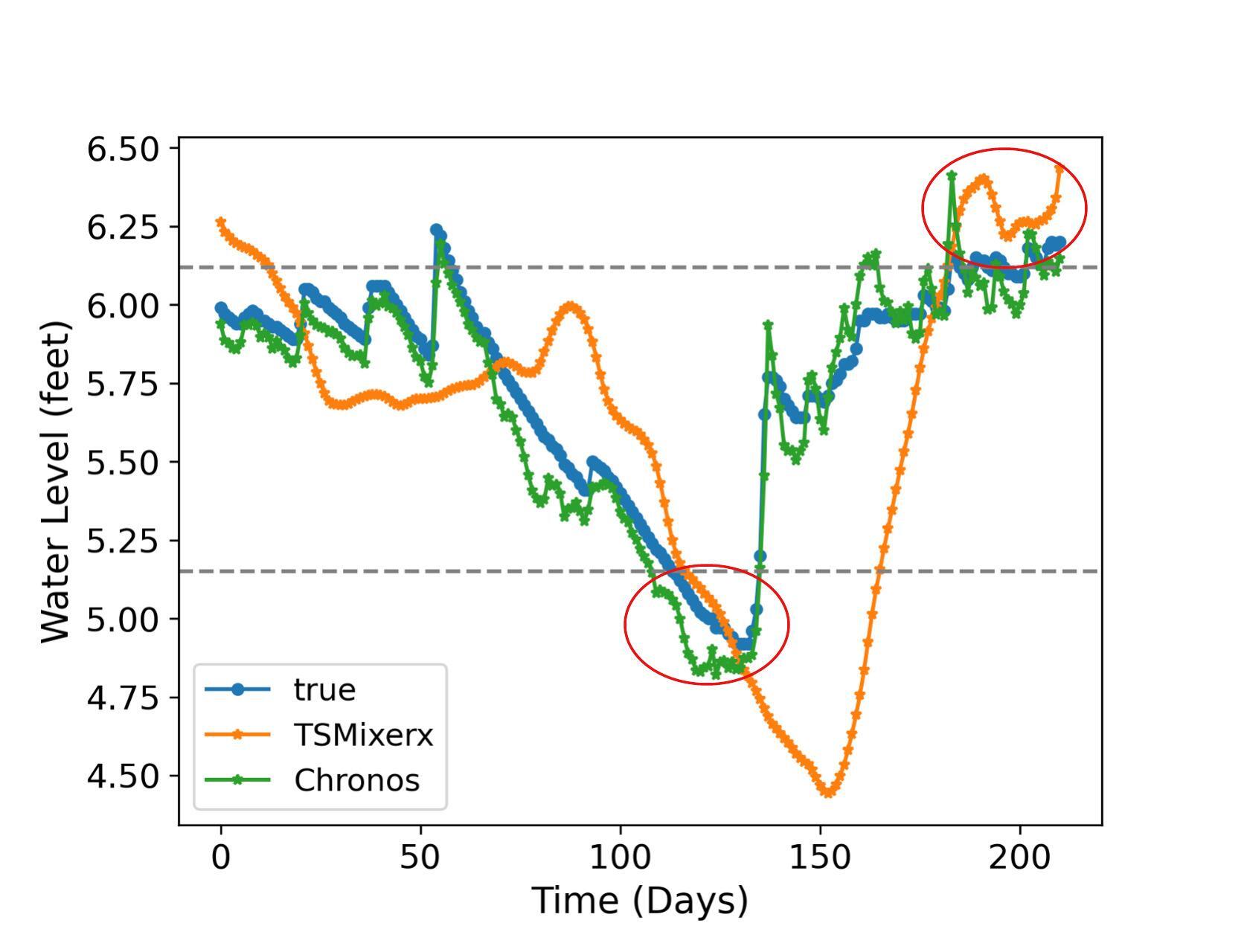}
        \caption{ }
        \label{fig:G620_visual}
    \end{subfigure}
    \hfill
    \begin{subfigure}[b]{0.33\textwidth}
        \includegraphics[width=\textwidth]{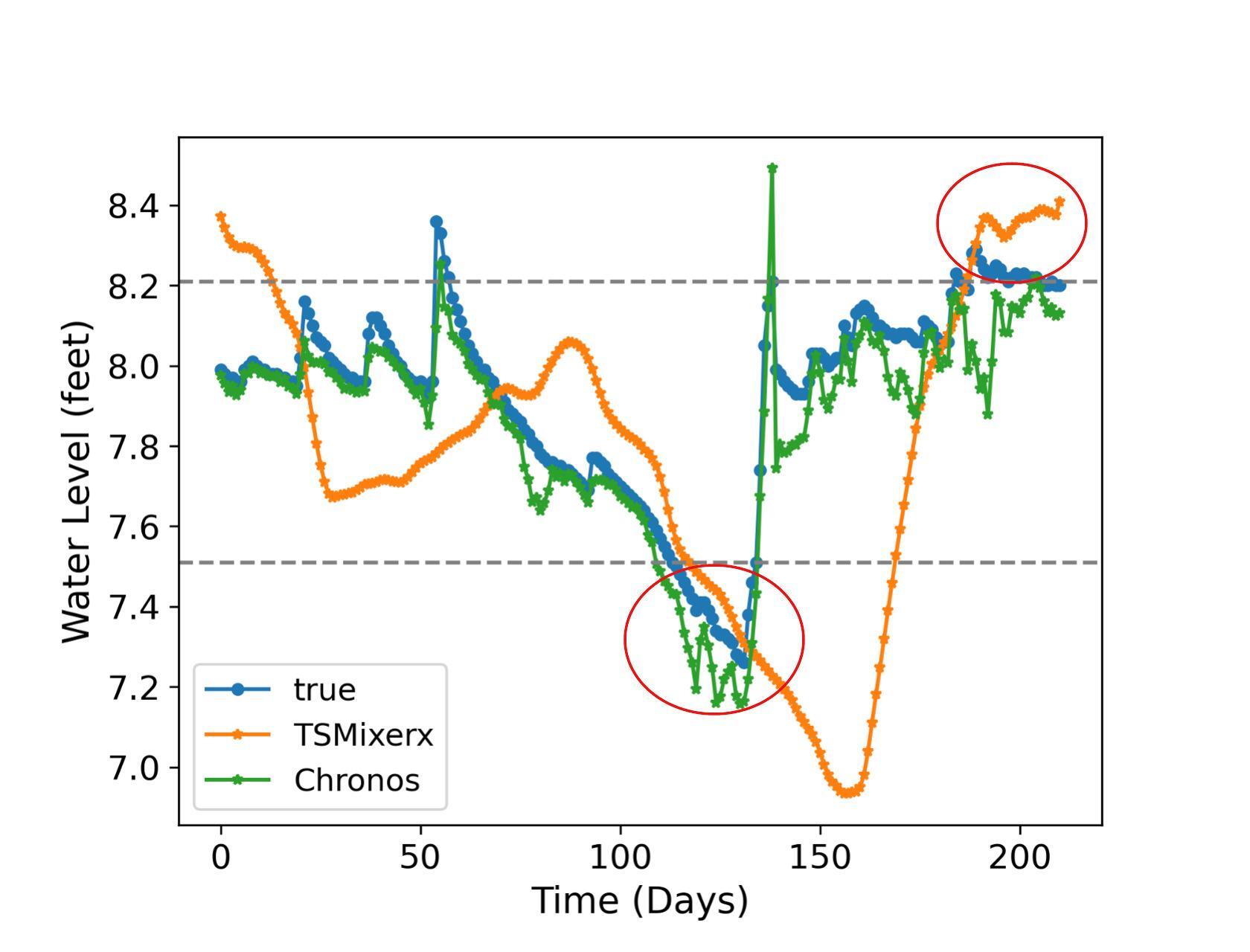}
        \caption{ }
        \label{fig:NESRS1_visual}
    \end{subfigure}

    \caption{Extreme value predictions at (a) P33, (b) G620, and (c) NESRS1 stations using \texttt{Chronos} and \texttt{TSMixerx} models.}
    \label{fig:three_stations}
\end{figure}

Figure \ref{fig:three_stations} highlights the extreme value predictions for the \texttt{Chronos} and \texttt{TSMixerx} models across the P33, G620, and NESRS1 measuring stations for a lead time of 28 days. The two horizontal lines in each plot represent the upper and lower thresholds for extreme values. Predictions that exceed these thresholds are circled, indicating instances where the models detected extreme water levels. \texttt{Chronos} consistently outperforms other models in extreme value prediction, while \texttt{TSMixerx}, despite its ordinary performance for overall water levels, successfully captures extreme events at these three stations. However, we observe false alarms in certain time periods, particularly between time steps 140 and 170. During this window, \texttt{TSMixerx} incorrectly predicts extreme water levels, which could potentially lead to unnecessary warnings or misinterpretations of water conditions.
%

\newpage
\section{Parameter Study for Input Length}
\label{sec:ParameterStudy_Chronos}
\begin{figure}[ht]
\centering
    \includegraphics[width=0.5\columnwidth]{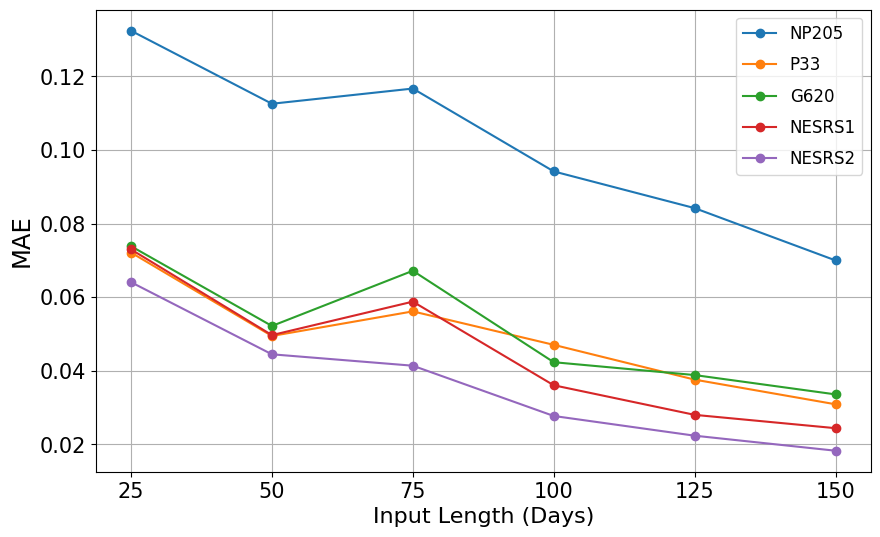}
\caption{Impact of Input Length on \texttt{Chronos} Model Performance at various stations.}
\label{fig:ParameterStudy_Chronos}
\end{figure}

As shown in Figure \ref{fig:ParameterStudy_Chronos}, we investigated how the \texttt{Chronos} model's predictive performance is affected by varying the input sequence length. Specifically, we tested input lengths of 25, 50, 75, 100, 125, and 150 days, and measured the resulting Mean Absolute Error (MAE) across five different measuring stations (NP205, P33, G620, NESRS1, and NESRS2).
The results show that as the number of input days increases, the model's performance improves, as evidenced by the decreasing MAE values.
This indicates that longer input sequences help the model make more accurate predictions. Interestingly, the NP205 measuring station exhibited slightly higher MAE values compared to the other stations, indicating a potential anomaly or unique characteristic of this station. 
On the other hand, the other stations (P33, G620, NESRS1, and NESRS2) followed a similar trend, with their MAEs closely clustered and showing a consistent improvement as the input sequence length grew.

\section{Correlation Study}
\label{sec:Correlation}
To investigate the discrepancy between the NP205 station and other water stations, we compute their correlations. The right plot emphasizes correlations among NP205, P33, G620, NESRS1, and NESRS2, while the left plot extends this analysis to include nearby gates. The results reveal that NP205 exhibits lower correlations with the other stations, with values ranging from 0.76 to 0.85, whereas the remaining stations maintain higher correlations (at least 0.85). 
This aligns with the spatial distribution shown in Figure \ref{fig:everglade_map}, where NP205 is positioned separately from the others. The hydrological flow dynamics, influenced by water management structures and natural flow patterns from WCA-3A and WCA-3B to Florida Bay, further explain these variations. 
\begin{figure}[ht!]
\centering
    \includegraphics[width=0.92\columnwidth]{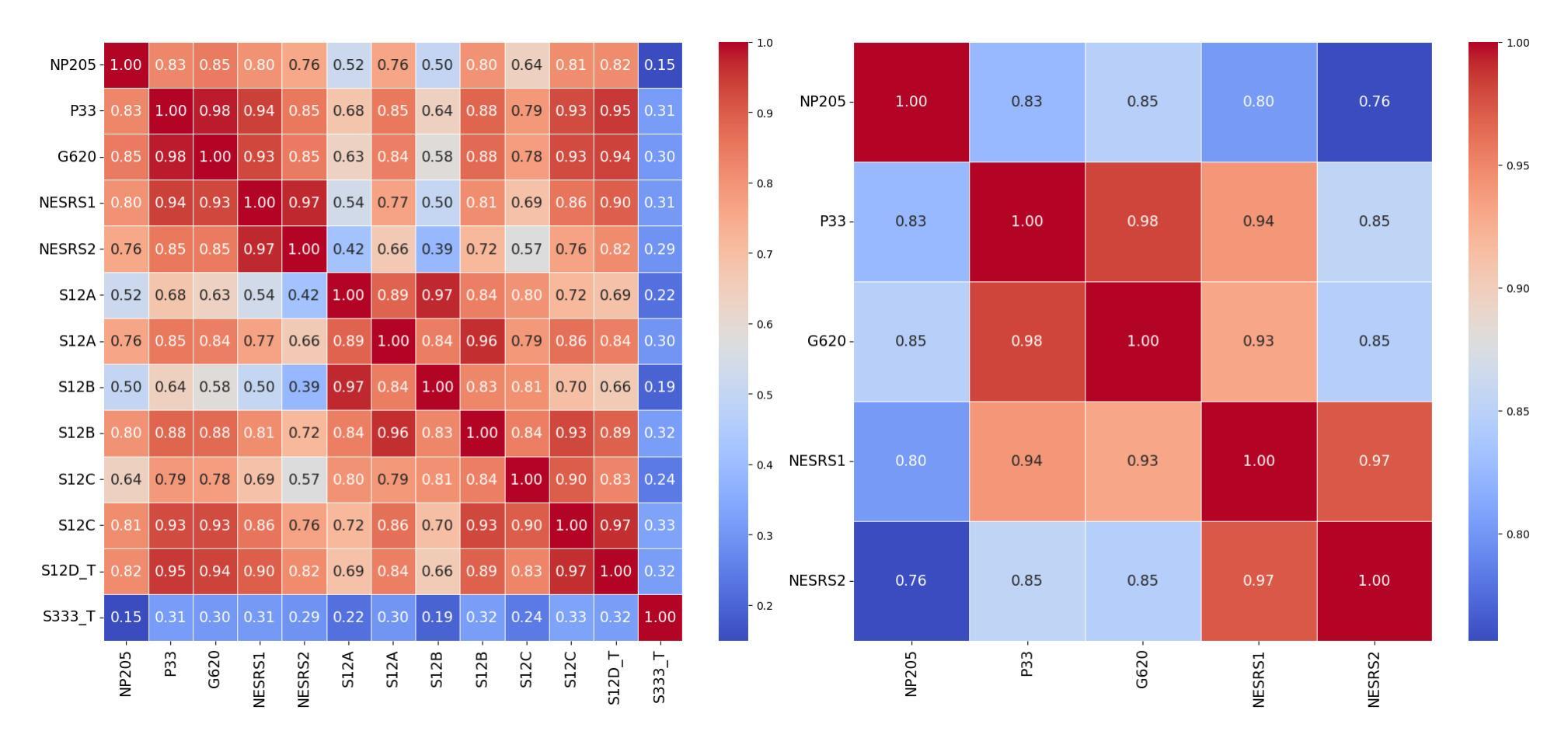}
\caption{Correlation Analysis of water levels at five measuring water stations and nearby gates.}
\label{fig:water_correlation}
\end{figure}


\newpage
\section{Visualization}
In the following, we provide the visualizations of the ground truth and prediction values of 16 models at different stations, while the corresponding MAE values are included in the figures.
\label{sec:more_visual}
\subsection{Details for the NP205 Station 
}
\label{sec:visualize_np205}
\begin{figure}[ht]
\centering
    \includegraphics[width=1\columnwidth]{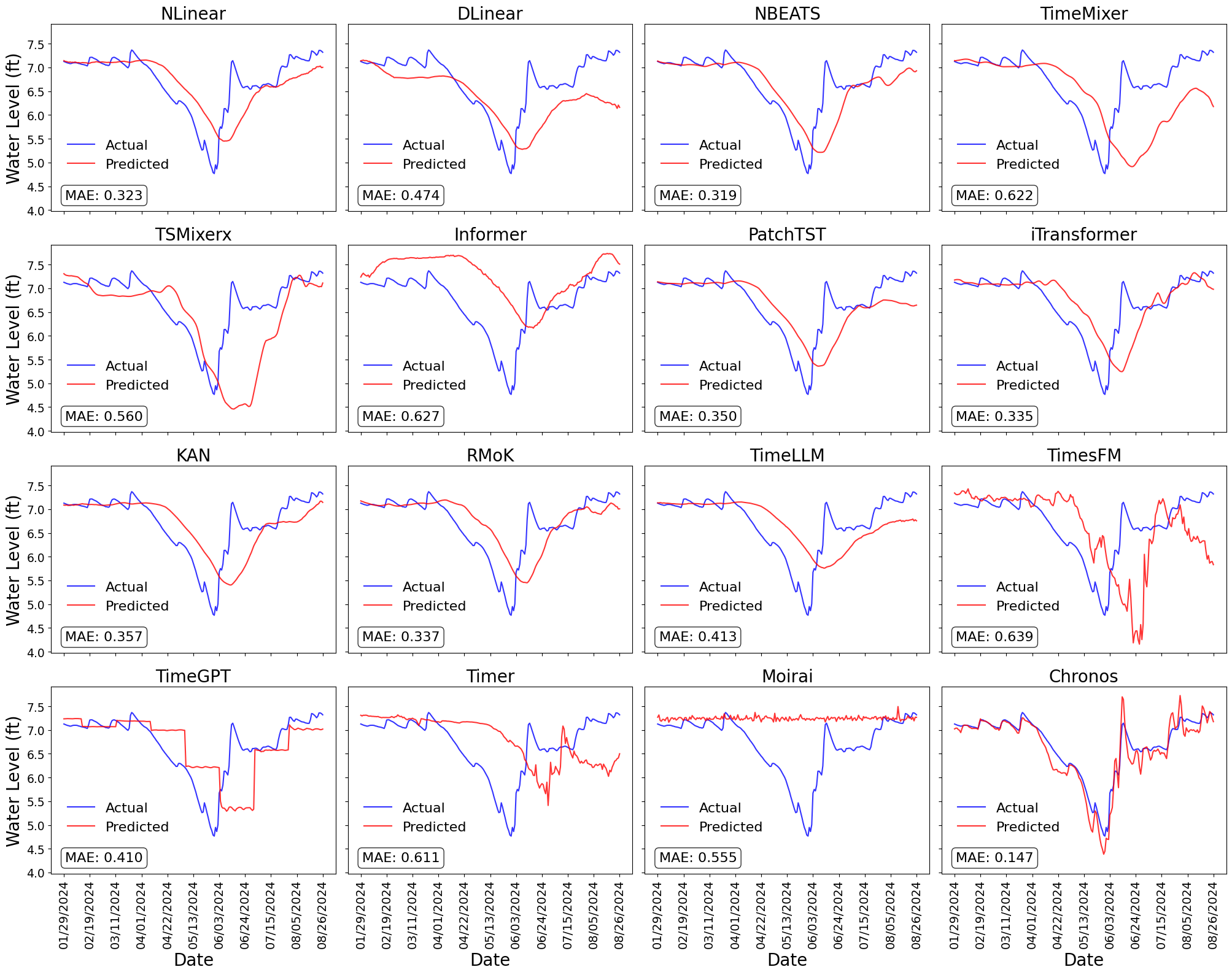}
\caption{Comparison of Model Performance for 28-Day Water Level Forecasting at NP205 Measuring Station.}
\label{fig:visualize_np205}
\end{figure}

The model comparison plots shown in Figure \ref{fig:visualize_np205} visualize the 28-day lead time water level forecasts at NP205, comparing actual observations with predictions from multiple machine learning models. Each subplot highlights predicted trends alongside actual values while also displaying the Mean Absolute Error (MAE), a key metric for evaluating predictive accuracy. The results indicate that \texttt{Chronos} consistently outperforms other models, achieving the lowest MAE of 0.147. This is consistent with the results shown in Table \ref{tab:overall_perform}, which comprehensively evaluates the performance of the models at multiple measuring stations. The analysis underscores the effectiveness of specialized time series models and pretrained foundation models in capturing hydrological patterns for improved water level forecasting.

\newpage
\subsection{Details for the P33 Station 
}
\label{sec:visualize_p33}
\begin{figure}[ht]
\centering
    \includegraphics[width=1\columnwidth]{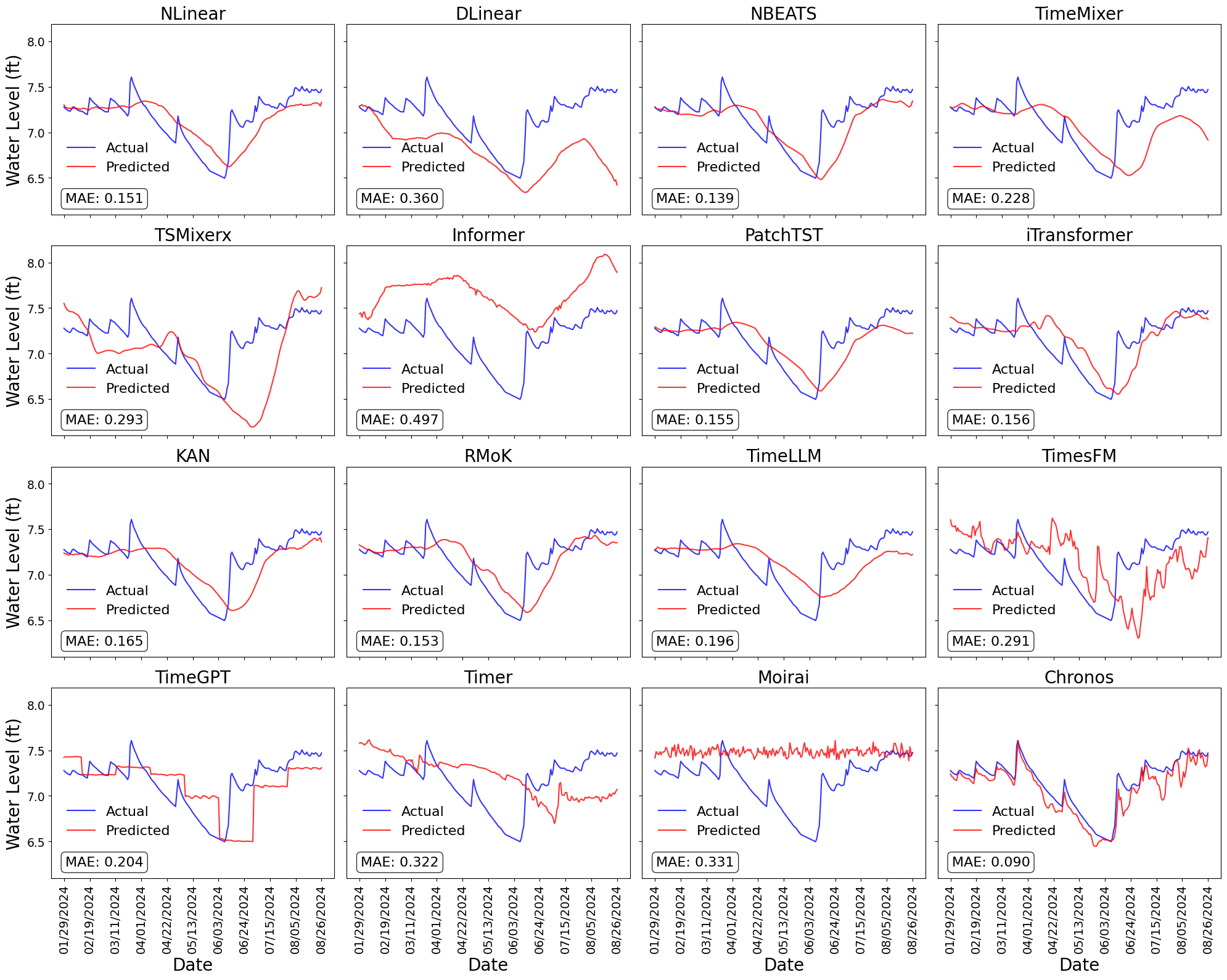}
\caption{Comparison of Model Performance for 28-Day Water Level Forecasting at P33 Measuring Station.}
\label{fig:visualize_p33}
\end{figure}

Next, the model comparison plots shown in Figure \ref{fig:visualize_p33} visualize the 28-day lead time water level forecasts at P33 Measuring Station, comparing actual observations with predictions from multiple models. 
Unlike NP205, where the models struggled with higher MAE values, all models performed better for the P33 measuring station, achieving lower MAE values across the board. The \texttt{Chronos} model again emerged as the most accurate, achieving the lowest MAE of 0.090, reinforcing its superior ability to capture hydrological patterns. In particular, the \texttt{NBEATS} model, which had an MAE of 0.319 for NP205,  improved significantly at P33, achieving a much lower MAE of 0.139. 

\newpage
\subsection{Details for the G620 Station 
}
\label{sec:visualize_g620}
\begin{figure}[ht]
\centering
    \includegraphics[width=1\columnwidth]{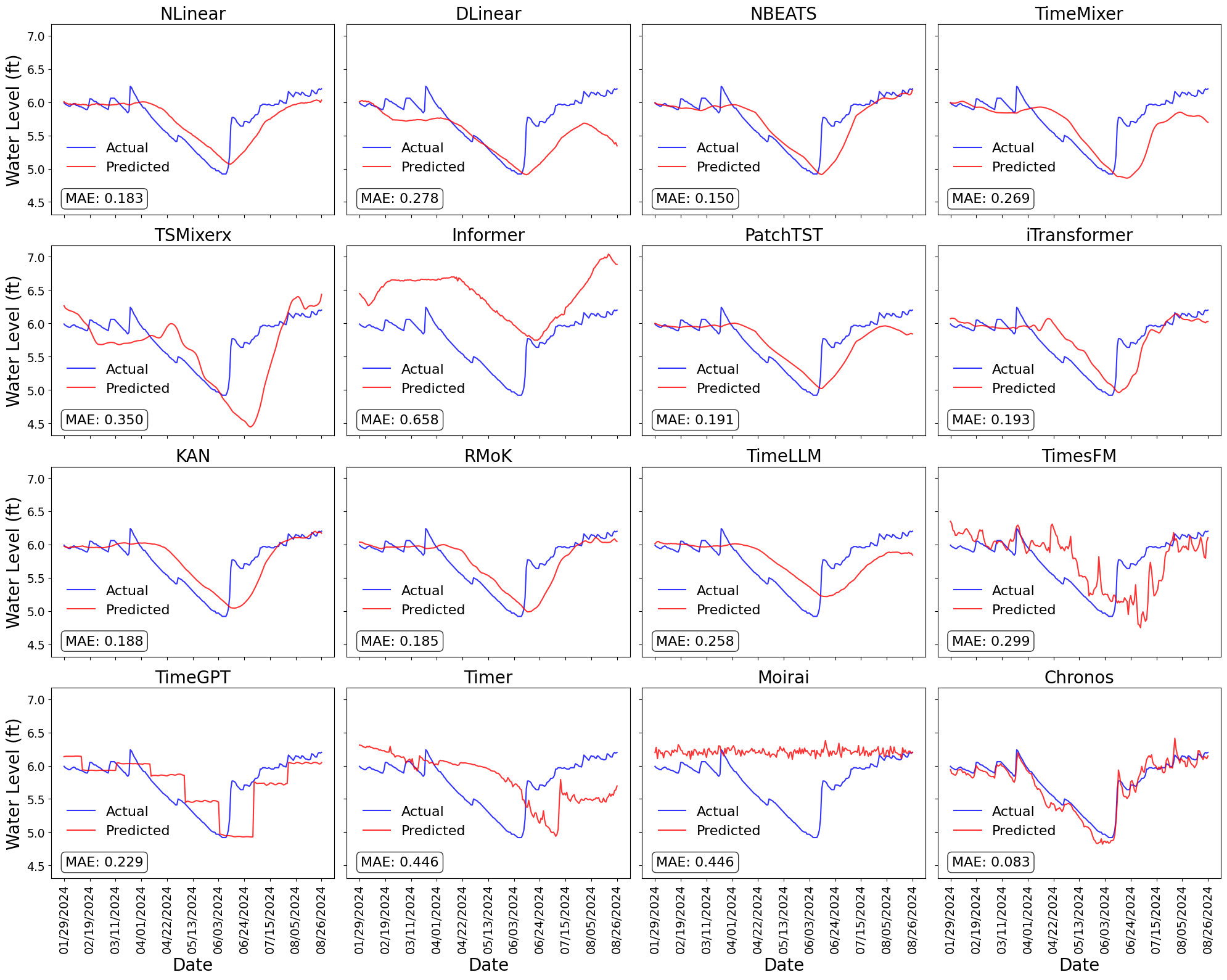}
\caption{Comparison of Model Performance for 28-Day Water Level Forecasting at G620 Measuring Station.}
\label{fig:visualize_g620}
\end{figure}

Figure \ref{fig:visualize_g620}  presents the 28-day lead time water level forecasts for G620, comparing the actual observations with predictions from multiple models. From the analysis, \texttt{Chronos} and \texttt{DLinear} performed notably better compared to their metrics for NP205 and P33. \texttt{Chronos} achieved the lowest MAE of 0.083, while \texttt{DLinear} improved significantly to 0.278 from its previous 0.360 at P33. However, other models exhibited slightly higher MAE values compared to P33. In particular, the \texttt{Informer} model showed a decline in performance, with its MAE increasing to 0.658 from 0.497 at P33, indicating a greater deviation of predicted values from actual observations.

\newpage
\subsection{Details for the NESRS1 Station 
}
\label{sec:visualize_nesrs1}
\begin{figure}[ht]
\centering
    \includegraphics[width=1\columnwidth]{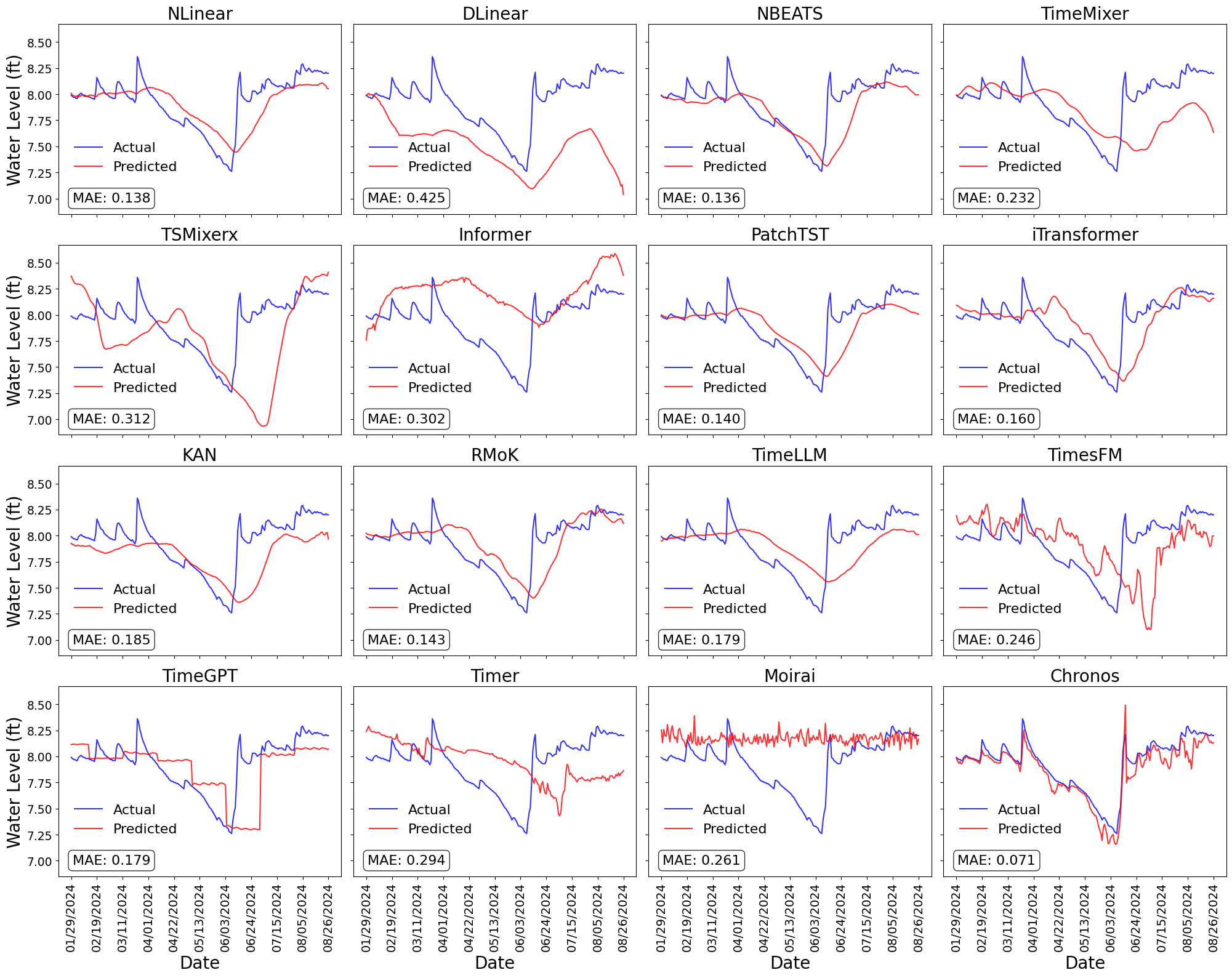}
\caption{Comparison of Model Performance for 28-Day Water Level Forecasting at NESRS1 Measuring Station.}
\label{fig:visualize_nesrs1}
\end{figure}

Figure \ref{fig:visualize_nesrs1} illustrates the 28-day lead-time water level forecasts at the NESRS1 measuring station, comparing actual observations with model predictions.  In all models, performance improved significantly compared to the measuring stations NP205, P33, and G620, with lower MAE values indicating improved predictive accuracy. However, an exception is observed with the \texttt{DLinear} model, which recorded an MAE of 0.425, higher than the 0.278 and 0.360 values at the G620 and P33 stations respectively. This suggests that while most models effectively capture water level patterns for NESRS1, \texttt{DLinear} struggles to maintain its prior performance. The \texttt{Chronos} model achieved the best performance with an MAE of 0.071, the lowest among NP205, P33, and G620 measuring stations, highlighting its effectiveness in capturing water level patterns at NESRS1. 

\newpage
\subsection{Details for the NESRS2 Station 
}
\label{sec:visualize_nesrs2}
\begin{figure}[ht]
\centering
    \includegraphics[width=1\columnwidth]{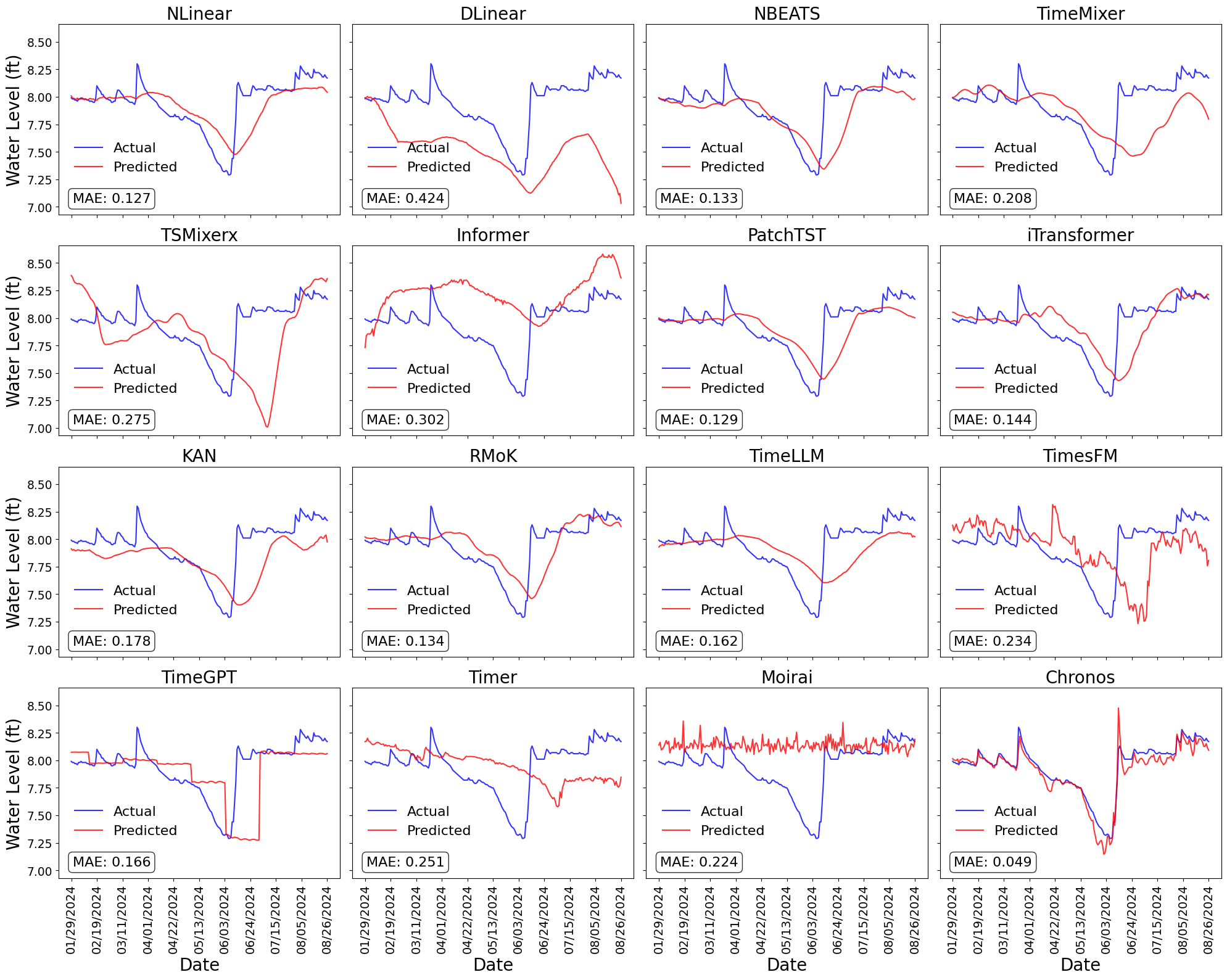}
\caption{Comparison of Model Performance for 28-Day Water Level Forecasting at NESRS2 Measuring Station.}
\label{fig:visualize_nesrs2}
\end{figure}

Figure \ref{fig:visualize_nesrs2} presents the 28-day lead time water level forecasts at NESRS2, demonstrating a consistent improvement in predictive accuracy across all models compared to the other measuring stations. The results indicate that NESRS2 exhibits the lowest overall MAE values, making it the most predictable among the five stations analyzed. As observed in prior analyses, the \texttt{Chronos} model continues to outperform others, achieving an MAE of 0.049, the lowest recorded across all stations. Additionally, \texttt{TSMixerx} improved significantly, reducing its MAE from 0.312 at NESRS1 to 0.275 at NESRS2. Similarly, \texttt{NLinear} also demonstrated a slight improvement, with its MAE decreasing from 0.138 to 0.127. These results highlight the effectiveness of specialized time series models in capturing hydrological patterns at NESRS2, further reinforcing the trend of improved predictions at this location.

\newpage
\section{Implementation Details}
The implementation of the forecasting models in this study spans a variety of architectures, ranging from linear models to deep learning-based transformers,large language models and foundation models. Table \ref{tab:imple_detail} provides a comprehensive comparison of key implementation parameters, including input size, learning rate, optimizer, library/API dependencies, architecture, training and inference strategies. Additionally, model complexity in terms of the number of parameters is depicted in Figure \ref{fig:ModelComplexity}.

\label{sec:imple_detail}
\begin{table*}[ht]
    \centering
    \resizebox{\textwidth}{!}{ 
    \begin{tabular}{l|c|c|c|c|c|c|c}
        \toprule
        \textbf{Model} & \textbf{Input Size} & \textbf{Learning Rate} & \textbf{Optimizer} & \textbf{Library/API} & \textbf{Architecture} & \textbf{Training} & \textbf{Inference}  \\
        \midrule
        NLinear & 100 & $1e^{-3}$  & Adam  & NeuralForecast\footnotemark[1] & Linear & Yes & zero-shot  \\
        DLinear & 100 & $1e^{-3}$ & Adam & NeuralForecast\footnotemark[1] & Linear  & Yes & zero-shot \\
        NBEATS & 100 & $1e^{-3}$ & Adam & NeuralForecast\footnotemark[1] & MLP & Yes & zero-shot \\
        TimeMixer & 100 & $1e^{-3}$ & Adam & NeuralForecast\footnotemark[1]  & MLP & Yes & zero-shot \\
        TSMixer & 100 & $1e^{-3}$ & Adam & NeuralForecast\footnotemark[1] & MLP & Yes & zero-shot  \\
        TSMixerx & 100 & $1e^{-3}$ & Adam & NeuralForecast\footnotemark[1] & MLP & Yes & zero-shot  \\
        Informer & 100 & $1e^{-3}$ & Adam & NeuralForecast\footnotemark[1] & Transformer & Yes & zero-shot \\
        PatchTST & 100 & $1e^{-3}$ & Adam & NeuralForecast\footnotemark[1] & encoder-only & Yes & zero-shot \\
        iTransformer & 100  & $1e^{-3}$ & Adam & NeuralForecast\footnotemark[1] & encoder-only & Yes & zero-shot \\
        KAN & 100 & $1e^{-3}$ & Adam & NeuralForecast\footnotemark[1] & KAN & Yes & zero-shot  \\
        RMok & 100 & $1e^{-3}$ & Adam & NeuralForecast\footnotemark[1] & KAN & Yes & zero-shot \\
        TimeLLM & 100 & $1e^{-3}$ & Adam & NeuralForecast\footnotemark[1] & LLM & Yes & zero-shot \\
        \midrule
        TimeGPT & 100 & $1e^{-4}$ & Adam & Nixtla API\footnotemark[2] & Transformer & No & zero-shot \\
        TimesFM & 128 & $1e^{-4}$ & Adam & HuggingFace\footnotemark[3] & decoder-only & No & zero-shot \\
        Timer & 100 & $1e^{-4}$ & Adam & HuggingFace\footnotemark[4]  & decoder-only & No & zero-shot \\
        Moirai & 100 & $1e^{-3}$ & Adam & HuggingFace\footnotemark[5] & encoder-only & No & zero-shot \\
        Chronos & 100 & $1e^{-5}$ & Adam & HuggingFace\footnotemark[6] & encoder-decoder & No & zero-shot \\
        \bottomrule        
    \end{tabular}
    }
    \caption{Implementation Details. The first 12 models are task-specific while the last 5 are pre-trained foundation models for time series modeling. }
    \label{tab:imple_detail}
\end{table*}
\footnotetext[1]{\url{https://github.com/Nixtla/neuralforecast}}
\footnotetext[2]{\url{https://docs.nixtla.io/docs/getting-started-about_timegpt}}
\footnotetext[3]{\url{https://huggingface.co/google/timesfm-1.0-200m-pytorch}}
\footnotetext[4]{\url{https://huggingface.co/thuml/timer-base-84m}}
\footnotetext[5]{\url{https://huggingface.co/Salesforce/moirai-1.0-R-large}}
\footnotetext[6]{\url{https://huggingface.co/autogluon/chronos-bolt-small}}

The NeuralForecast library, part of the Nixtlaverse ecosystem, provides a scalable and user-friendly framework for implementing deep learning models for time series forecasting. In this study, several models from this library were utilized, including \texttt{NLinear}, \texttt{DLinear}, \texttt{NBEATS}, \texttt{TimeMixer}, \texttt{TSMixer}, \texttt{TSMixerx}, \texttt{Informer},\texttt{PatchTST}, \texttt{iTransformer}, \texttt{KAN}, \texttt{RMok} and \texttt{TimeLLM}, as listed in Table \ref{tab:imple_detail}. NeuralForecast requires input data in a long-format structure, where each time series is stored as individual rows in a Pandas DataFrame, with corresponding timestamps and values. This format ensures compatibility across multiple time series and facilitates batch processing for neural network-based models. 
\texttt{TimeGPT}, which is Nixtla’s generative pre-trained transformer for time series forecasting, was evaluated using the Nixtla API, enabling seamless model inference without requiring fine-tuning.

While most models in Table \ref{tab:imple_detail} use an input size of 100, \texttt{TimesFM} is assigned an input length of 128, deviating from the standard. The architectural constraint of \texttt{TimesFM} requires input lengths to be multiples of 32. For this study, the TimesFM-1.0-200M checkpoint (timesfm-1.0-200m-pytorch) was used, which supports univariate time series forecasting with context length up to 512. The \texttt{Timer} model was implemented using the Hugging Face transformers library, leveraging its pre-trained Timer-base-84M checkpoint. This model follows a Causal Transformer (Decoder-only) architecture and is loaded using the AutoModelForCausalLM class, enabling generative forecasting for time-series data. Salesforce's Uni2TS, a PyTorch-based library was used to evaluate the \texttt{Moirai} model for time series forecasting. Inference was conducted in a zero-shot setting without fine-tuning, leveraging the GluonTS framework. The \texttt{Chronos} model was implemented using the AutoGluon-TimeSeries (AG-TS) library, which provides an easy-to-use framework for time series forecasting. The model was accessed through the TimeSeriesPredictor API, specifying the "amazon/chronos-bolt-small" checkpoint for inference.





\end{document}